\documentclass{vgtc}                          % final (conference style)
\usepackage{amsmath}
\usepackage{amssymb}
\usepackage{subcaption}
\usepackage{soul}

\graphicspath{{figures/}{pictures/}{images/}{./}} % where to search for the images

\usepackage{times}                     % we use Times as the main font
\usepackage{mathptmx}                  % use matching math font

\onlineid{0}

\vgtccategory{Research}

\vgtcinsertpkg

\author{Jixian Li\thanks{e-mail: jixianli@sci.utah.edu}\\ %
        \scriptsize SCI Institute %
\and Timbwaoga Aime \\Judicael Ouermi\thanks{e-mail: touermi@sci.utah.edu}\\ %
     \scriptsize SCI Institute %
\and Mengjiao Han\thanks{e-mail: hanm@anl.gov}\\ %
     \scriptsize Argonne National Laboratory
\and Chris R. Johnson\thanks{e-mail: crj@sci.utah.edu}\\
    \scriptsize SCI Institute %
    }

\title {Uncertainty Tube Visualization of Particle Trajectories}
\abstract{
Predicting particle trajectories with neural networks (NNs) has substantially enhanced many scientific and engineering domains. However, effectively quantifying and visualizing the inherent uncertainty in predictions remains challenging. Without an understanding of the uncertainty, the reliability of NN models in applications where trustworthiness is paramount is significantly compromised. This paper introduces the \textit{uncertainty tube}, a novel, computationally efficient visualization method designed to represent this uncertainty in NN-derived particle paths. Our key innovation is the design and implementation of a superelliptical tube that accurately captures and intuitively conveys nonsymmetric uncertainty. By integrating well-established uncertainty quantification techniques, such as Deep Ensembles, Monte Carlo Dropout (MC Dropout), and Stochastic Weight Averaging-Gaussian (SWAG), we demonstrate the practical utility of the \textit{uncertainty tube}, showcasing its application on both synthetic and simulation datasets. 
} 


\def\change#1{{#1}}

\newcommand{\synth}{\textbf{synth}}
\newcommand{\swaglr}{swag\_lr}
\newcommand{\nswagsamples}{n\_swag\_samples}
\newcommand{\sgdweightdecay}{sgd\_weight\_decay}
\newcommand{\sgdmomentum}{sgd\_momemtum}
\newcommand{\halfcylinder}{\textbf{half cylinder}}

\begin{document}

% uncomment for using teaser
\teaser{
 \includegraphics[width=0.99\linewidth]{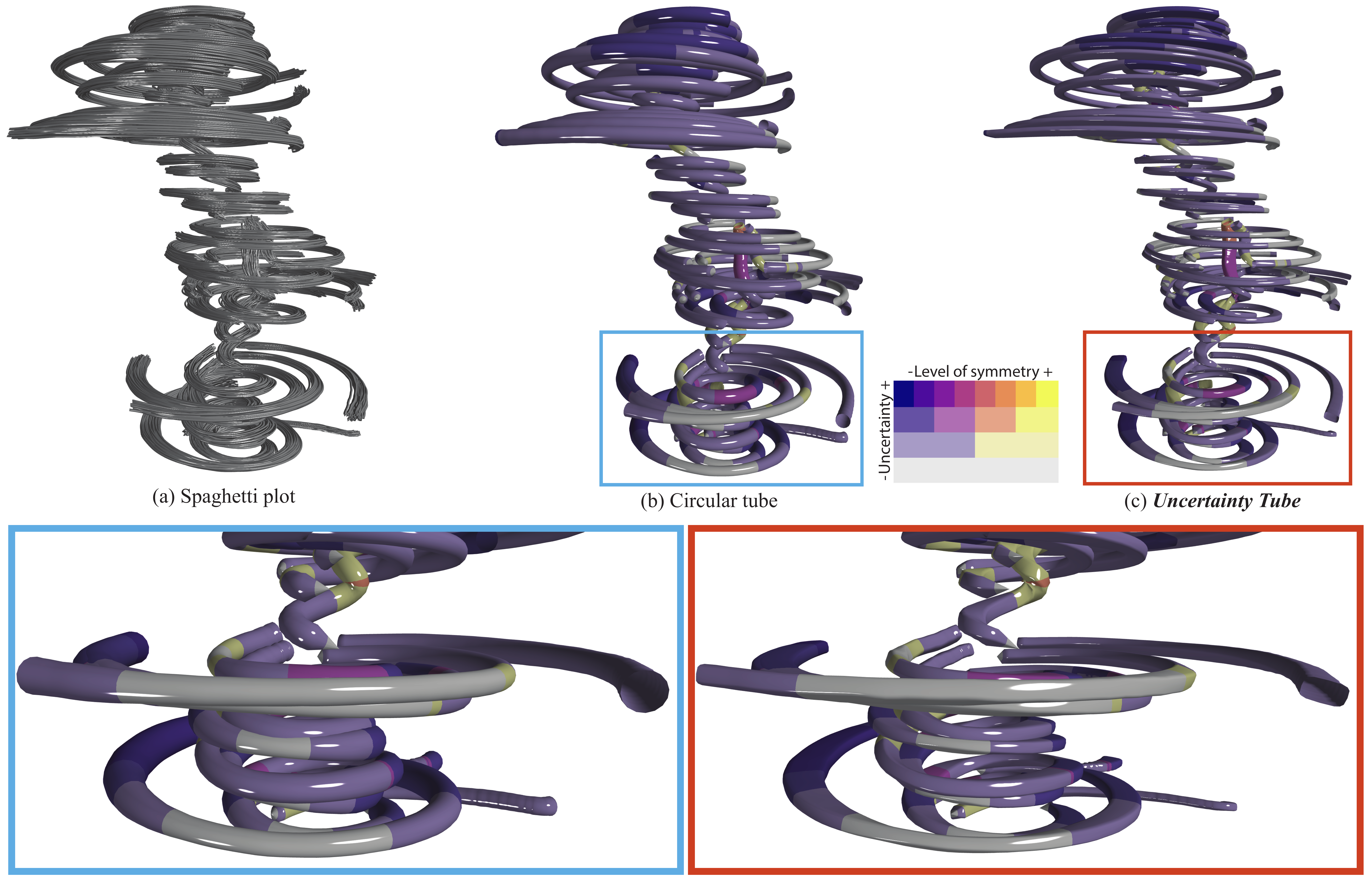}
 \centering
 \caption{Comparison of flow map uncertainty visualization techniques for the \textbf{tornado} dataset. This figure compares (a) a spaghetti plot of ensemble members, (b) a circular tube, and (c) our \textit{uncertainty tube} for visualizing model uncertainty. Previous methods face challenges such as visual clutter (a) or the assumption of symmetric uncertainty (a, b), but our \textit{uncertainty tube} (c), constructed using superellipses, provides a more accurate visualization of asymmetric uncertainty. Its superelliptical shape distinctly improves the visualization of the uncertainty orientation and its evolution along trajectories, as highlighted in the boxes. The visualization is further enhanced with a color palette that uses gray for low uncertainty, blue for large asymmetric uncertainty, and yellow for large symmetric uncertainty.} 
 %\textcolor{orange}{Mengjiao: add explanation of the color encoding in the visualization, as the colormap can be confusing.}
\label{fig:teaser}
}
\firstsection{Introduction}
\maketitle
%-------------------------------------------------------------------------

%-------------------------------------------------------------------------

Understanding and analyzing flow field data is fundamental for numerous scientific and engineering disciplines, including fluid dynamics, atmospheric science, and material processing. Traditional computational fluid dynamics (CFD) simulations are often computationally intensive, a limitation that has led researchers to explore more efficient paradigms. This exploration has given rise to neural networks (NNs) as a transformative tool in this domain, driven by their capacity to overcome these computational bottlenecks. NNs are now widely employed for tasks such as learning and emulating complex turbulent fluid dynamics, enabling the rapid reconstruction of intricate flow fields from limited data, performing super-resolution on coarse simulation outputs, and serving as highly efficient surrogate models that can predict flow behavior orders of magnitude faster than conventional methods~\cite{mario2023,Pranshu2021,Nils2020,CALZOLARI2021108315}. Notably, recent work, such as Han et al.~\cite{han2024interactive,han2021exploratory}, leverages NNs to learn Lagrangian-based flow maps, enabling efficient and robust particle tracing in time-varying fields. These data-driven models demonstrate remarkable accuracy and speed, making them increasingly indispensable for accelerating discovery and design cycles in fluid dynamics.

Despite these advancements, a significant challenge remains in providing a comprehensive understanding of the confidence associated with NN predictions in flow fields. Although NNs can effectively capture complex dynamics, their outputs are typically deterministic and lack explicit representations of predictive uncertainty. To address this limitation, several uncertainty quantification (UQ) methods have been developed for NNs, aiming to provide measures of prediction reliability ~\cite{ABDAR2021243,GANAIE2022105151,Rahaman2021,gal16dropout,maddox_2019_swag,lakshminarayanan2017simple}. These techniques aim to estimate the confidence a model has in its output, distinguishing between inherent data variability and the model's uncertainty due to limited knowledge. However, effectively communicating these complex, often multidimensional uncertainty estimates to researchers, particularly for dynamic elements such as particle trajectories, remains a key challenge.

% \todo{rewrite from here}
% Current research in uncertainty visualization frequently employs techniques such as error bars, confidence intervals, or probability distributions to indicate variability. 
% For particle paths uncertainty visualization, common approaches include drawing multiple perturbed trajectories, using glyphs to represent local uncertainty, encoding uncertainty through visual attributes, such as opacity, blur, or color saturation along the path.
A recent work in uncertainty-aware deep neural representation of vector fields by Kumar et al.~\cite{Kumar2025_vectorfield_uncertainty}, which is closely related to our problem, utilizes circular tubes with varying radii to represent the variation in steamlines. While the circular tube provides valuable insights into trajectory variability, it assumes symmetric uncertainty bounds and can oversimplify and misrepresent scenarios where uncertainty is inherently asymmetric. %Effectively visualizing such asymmetric uncertainty in particle trajectories, particularly in an intuitive and computationally efficient manner, remains an active area of research.

To address this limitation, this paper introduces the \textit{uncertainty tube}: a novel, computationally efficient visualization method designed to represent prediction uncertainty in (NN)-derived particle paths. \change{We focus our use case on neural network-based trajectories, but our method is generalizable to ensembles from other sources, such as simulations. This \textit{uncertainty tube} is designed to accurately capture nonsymmetric uncertainty distributions, highlighting the direction of variations.} The contribution of the paper includes:
\begin{enumerate}
    \item The uncertain estimation from ensemble trajectories and the design of the \textit{uncertainty tube} using superellipses and color palettes for accurate and enhanced visualization.
    \item The use of three datasets (\textbf{synth, tornado, half cylinder}) to compare the uncertainty quantification methods (Monte Carlo dropout, Deep Ensembles, and SWAG) and demonstrate the capability of the \textit{uncertainty tube} to accurately convey uncertainty in particle trajectories. 
\end{enumerate}

%-------------------------------------------------------------------------

\section{Related Works}
\subsection{Flow Field Visualization}
\subsubsection{Lagrangian Flow Reconstruction and Visualization}
% Visualizing and analyzing time-varying vector fields remains challenging due to the complexity of temporal evolution and the need to accurately capture dynamic flow characteristics.
% %
% A fundamental approach is particle integration, where seed points are advected through the vector field to generate pathlines that reveal flow behavior over time~\cite{mcloughlin2010over}.
%
Eulerian and Lagrangian reference frames are commonly used to represent time-varying flow fields.
%
% In an Eulerian representation, velocity fields are stored, and particle trajectories are computed by integrating these fields over time.
% %
% In contrast, a Lagrangian representation encodes flow behavior using flow maps, which store the start and end positions of particles from time $t_0$ to $t_1$, and allow the computation of arbitrary trajectories via interpolation.
In the Eulerian representation, velocity fields are stored, and particle paths are computed by integrating over time.
In contrast, the Lagrangian representation uses flow maps to store particle start and end positions over a given time interval, enabling trajectory computation through interpolation.
Each approach has its advantages and limitations.
The Eulerian method is computationally efficient but requires a dense temporal resolution to achieve accurate trajectory reconstruction~\cite{da2004lagrangian,qin2014quantification,agranovsky2014improved,sane2018revisiting,rockwood2019practical,sane2021investigating}.
The Lagrangian method offers a good accuracy–storage trade-off for exploring temporally sparse datasets~\cite{agranovsky2014improved, rapp2019void,sane2021investigating,sane2022exploratory} and directly supports feature extraction~\cite{froyland2015rough,schlueter2017coherent,hadjighasem2017critical,froyland2018robust,Jakob2020}. As a result, it has received increasing attention in recent years.

In Lagrangian representation, the flow maps are typically computed in situ, while particle trajectories are reconstructed post hoc through interpolation.
The accurate and fast reconstruction of new trajectories from the flow maps is an important component of post hoc analysis. 
%
% Multiple methods have been proposed to improve particle trajectory, such as interpolation through multiresolution refinement~\cite{agranovsky2015multi}, parametric curve representations~\cite{bujack2015lagrangian}, and efficient neighborhood search with k-d trees~\cite{chandler2014interpolation}.
\change{Several methods have been proposed to enhance reconstruction accuracy and accelerate neighbor lookup in particle trajectory reconstruction, including multiresolution refinement through interpolation~\cite{agranovsky2015multi}, parametric curve representations~\cite{bujack2015lagrangian}, and efficient neighborhood search using k-d trees~\cite{chandler2014interpolation}.}
%
% Agranovsky et al. presented a multiresolution interpolation scheme that begins with a
% base resolution and adds additional trajectories if the region contains
% interesting behaviors~\cite{agranovsky2015multi}. 
% %
% Bujack et al.~\cite{bujack2015lagrangian} proposed using parametric curves to represent particle trajectories, such as Bézier curves and Hermite splines, to improve the aesthetics of the derived trajectories.
% %
% Chandler et al.~\cite{chandler2014interpolation} developed a k-d tree to enable efficient lookup of particle neighborhoods during interpolation.
%
However, these methods still face challenges in supporting real-time interpolation and visualization of flows.
%
%TAJO Two main challenges of interactive post hoc analysis are the high I/O overhead of loading high-resolution flow maps and the computational cost of accelerating cell lookups for particle neighborhoods.
%
% Additionally, unstructured flow maps require time-consuming triangulation or tetrahedralization, further slowing the interpolation process.
% 
To address these challenges, scientists have been exploring deep-learning-based approaches~\cite{han2021exploratory,han2024interactive}, which will be described in the next subsection.

\subsubsection{Deep Learning for Flow Visualization}
In recent years, deep learning has gained significant traction in the field of flow visualization~\cite{liu2022deep}.
They have been applied to a wide range of tasks, such as optimizing data access patterns to improve performance in distributed memory particle advection~\cite{hong2018access} and segmenting streamlines~\cite{li2015extracting}.
%
% detecting eddies and vortices~\cite{lguensat2018eddynet,duo2019oceanic,franz2018ocean,tatarenkova2020edge,zhang2025vortextransformer}, extracting stable reference frames from unsteady 2D vector fields~\cite{kim2019robust}, 
% yi2018cnn, strofer2018data, bai2019streampath,liu2019cnn,deng2019cnn, wang2021rapid, kashir2021application,beck2020neural,
Deep learning has also been leveraged for selecting representative sets of particle trajectories~\cite{sane2020survey}, often using clustering approaches informed by learned features~\cite{han2018flownet, lee2021deep}. \change{In addition, many of the deep-learning techniques have been extended to directly integrate physical and conservation laws into the NN ~\cite{RAISSI2019686,arzani2022machine,pmlr-v119-sanchez-gonzalez20a}. These physics-informed deep-learning approaches allow the models to learn from data while respecting underlying physical principles, leading to improved robustness and accuracy.}
% pfaff2021learning is GNN not PINN
%
Given the scale of flow datasets, data reduction and reconstruction have become prominent areas of focus.
Several studies have demonstrated the use of low-resolution vector fields~\cite{guo2020ssr,gao2021super,hohlein2020comparative} or 3D streamlines~\cite{han2019flow, sahoo2021integration} to reconstruct high-resolution flow fields. 
In addition, recent work has explored temporal super-resolution of time-varying vector fields~\cite{han2022tsr,bao2023deep}.
Jakob et al.~\cite{Jakob2020}, for example, upsampled 2D FTLE scalar fields derived from Lagrangian flow maps by employing efficient super-resolution architectures.
% such as the ESPCN~\cite{shi2016real} and SRCNN~\cite{dong2015image}.
%
More recently, Sahoo et al.~\cite{Sahoo2022} introduced a method for compressing and reconstructing time-varying flow fields using implicit neural representations, demonstrating the promise of NNs for scalable and accurate flow reconstruction.

Effectively visualizing flow map data depends on two key factors: (1) accurately reconstructing particle trajectories and (2) enabling interactive visualization and exploration of these trajectories, as discussed in the previous section.
Han et al.~\cite{han2021exploratory} were the first to employ a multilayer perceptron (MLP) architecture to reconstruct Lagrangian-based flow maps for a 2D analytical dataset.
%
% While they demonstrated the feasibility and accuracy of using neural networks for this task, their initial study did not include comprehensive quantitative and qualitative evaluations across diverse datasets, nor did it explore potential modifications to the model architecture.
% %
% In addition, they did not investigate how the rapid inference capability of deep learning models could support interactive visualization.
Their follow-up work~\cite{han2024interactive} extended this by validating the approach on diverse 2D and 3D datasets and introducing a web-based viewer for interactive flow visualization using fast neural inference.
% They showed that MLP-based models can accurately reconstruct particle trajectories for both 2D and 3D flows, whether the underlying data is structured or unstructured.
%
% TAJO Moreover, by leveraging the fast inference speed of NNs, they developed a web-based viewer that enables interactive post hoc analysis and visualization of time-varying flow fields.
%
While deep learning has been increasingly applied to flow visualization, existing works rarely explore model uncertainty in detail.
In this work, we build upon the model proposed by Han et al.~\cite{han2024interactive} and extend it with a specific focus on evaluating model uncertainty.

% To address these limitations, Han et al.~\cite{han2024interactive} subsequently provided a more in-depth study of the performance.
%

\subsection{Estimating Uncertainty in Machine Learning Models}
Addressing the uncertainty in deep learning is an active research area, with various approaches offering different trade-offs in rigor, cost, and performance. In this paper, we focus on the model's epistemic uncertainty instead of aleatoric uncertainty resulting from noise and randomness in the data. Here, we list some of the common uncertainty estimation methods:

\textbf{Deep Ensembles \cite{lakshminarayanan2017simple}}: This method involves training multiple NNs independently from different random initializations. The variance of predictions across these ensemble members quantifies uncertainty. Deep Ensembles are empirically robust and well-calibrated, but incur high computational costs during training due to the requirement of multiple full models.

\textbf{Monte Carlo Dropout (MC Dropout) \cite{gal16dropout}}: MC Dropout adapts dropout regularization for uncertainty estimation during inference. This approach is computationally efficient because it does not require additional training if dropout is already being used. 
Despite these benefits and its wide adoption, others argue that its predictive distribution does not align with a true Bayesian posterior~\cite{folgoc2021mc}. Still, MC Dropout remains a valuable technique for capturing model uncertainty in many applications.

\textbf{Bayesian Neural Networks (BNNs) and Approximations}: BNNs aim to model a probability distribution over network weights, directly accounting for epistemic uncertainty. However, the exact Bayesian inference in NNs is currently intractable due to the high dimensionality and complexity of the posterior distribution. 
%TAJO To address this limitation, various approximation methods have been developed:
To address this limitation, various approximation methods such as Variational Inference~\cite{blundell2015weight}, Markov Chain Monte Carlo~\cite{welling2011bayesian}, Laplace Approximation~\cite {mackay1992practical}, and Stochastic Weight Averaging-Gaussian (SWAG) \cite{maddox_2019_swag}. 
%In this paper, we utilized Deep Ensembles, MC Dropout, and SWAG to estimate our model's uncertainty as they are commonly used in practice.

\change{In this paper, we utilized the two methods used in Kumar et al.~\cite{Kumar2025_vectorfield_uncertainty}, the Deep Ensembles and MC Dropout. We also added SWAG to demonstrate how our visualization can be applied to different ensemble-based methods. Both MC Dropout and SWAG introduce minimal computational overhead, which allows interactive exploration and visualization of flow map uncertainty.}

\subsection{Uncertainty Visualization}

Effectively visualizing uncertainty is important for data analysis and decision-making ~\cite{dong2012Uncertainty,reyes2025trusting}. Researchers have developed numerous visualization techniques to communicate uncertainty, ranging from fundamental statistical glyphs to more advanced probabilistic representations~\cite{hansen2014scientific}. Notable contributions in the field include the early review by Pang et al.~\cite{Pang1997approaches}, discussions on challenges and approaches by Johnson and Sanderson \cite{Johnson2003}, and the comprehensive taxonomy of uncertainty visualization approaches by Potter et al.~\cite{potter2012taxonomy}. More recently, Kamal et al.~\cite{Kamal2021recent} provided a survey on recent advances and ongoing challenges in the field, highlighting the continuous importance of depicting data quality and variability to ensure accurate interpretation.

Vector and flow field uncertainty visualization is challenging due to the directional and dynamic nature of these fields. These challenges are further exacerbated when considering time-varying, multiple computational fields, and large-scale data. To convey uncertainty, most techniques focus on representing variability of both direction and magnitude. Early work explored glyph-based approaches, such as those by Wittenbrink et al.~\cite{Wittenbrink1996glyphs}, for visualizing uncertainty in vector fields. More recently, Ouermi et al.~\cite{Ouermi2024glyph} have advanced glyph-based uncertainty visualization for time-varying vector fields. To represent topological properties of uncertain fields, Otto et al.~\cite{Otto20102dtopology} and Otto et al.~\cite{Otto20113dtopology} introduced methods for uncertain 2D and 3D vector field topology. 
%TAJO A challenging but crucial aspect is visualizing ensemble trajectories, which represent multiple possible paths or outcomes from simulations. For these ensembles, 
Mirzargar et al.~\cite{Mirzargar2014curve} proposed curve boxplots as a generalization of traditional boxplots for ensembles of curves, providing statistical summaries of trajectory bundles. In addition to the depth-based boxplot idea, Ferstl et al.~\cite{Ferstl2016streamline} introduced variability plots to cluster and characterize the major trends in the ensemble. Kumar et al.~\cite{Kumar2025_vectorfield_uncertainty} have explored uncertainty-aware deep neural representations to aid in the visual analysis of vector field data. 
\change{Extending Kumar et al.\cite{Kumar2025_vectorfield_uncertainty}, which uses circular tubes with varying radii to represent uncertainty of integralines, we use the \textit{uncertainty tube} to better represent the non-symmetric uncertainty.}

\section{Background}

\subsection{Deep-Learning-Based Largrangian Flow Maps}
Han et al.~\cite{han2021exploratory} introduced the first multilayer perceptron (MLP)-based model to explore time-varying vector fields using Lagrangian-based flow maps, which they later improved for more accurate predictions and evaluated on multiple datasets~\cite{han2024interactive}. In our paper, the uncertainty measurements are based on the flow map NN proposed by Han et al.~\cite{han2024interactive}, which is described below.

% Han et al.~\cite{han2021exploratory} introduced the first multilayer perceptron (MLP)-based model to explore time-varying vector fields using Lagrangian-based flow maps. 
% %
% Their workflow begins with in situ processing to extract these flow maps, followed by training deep neural networks on the extracted data to learn the behavior of the flow field.
% %
% To demonstrate and evaluate their method, they first conducted a detailed performance study using the well-known analytical dataset, the Double Gyre. 
% %
% Building on this initial work, they later improved the model by incorporating a sine activation function, expanding the evaluation to include multiple datasets, and developing a web-based viewer that allows interactive visualization of particle trajectories using the trained model~\cite{han2024interactive}.
% %
% \textcolor{orange}{In our paper, uncertainty measurements are based on the flow map neural network proposed by Han et al.~\cite{han2024interactive}. 
% %
% This section summarizes their model as a foundation for our approach.}

\begin{figure}[!ht]
    \centering
    \includegraphics[width=0.99\linewidth]{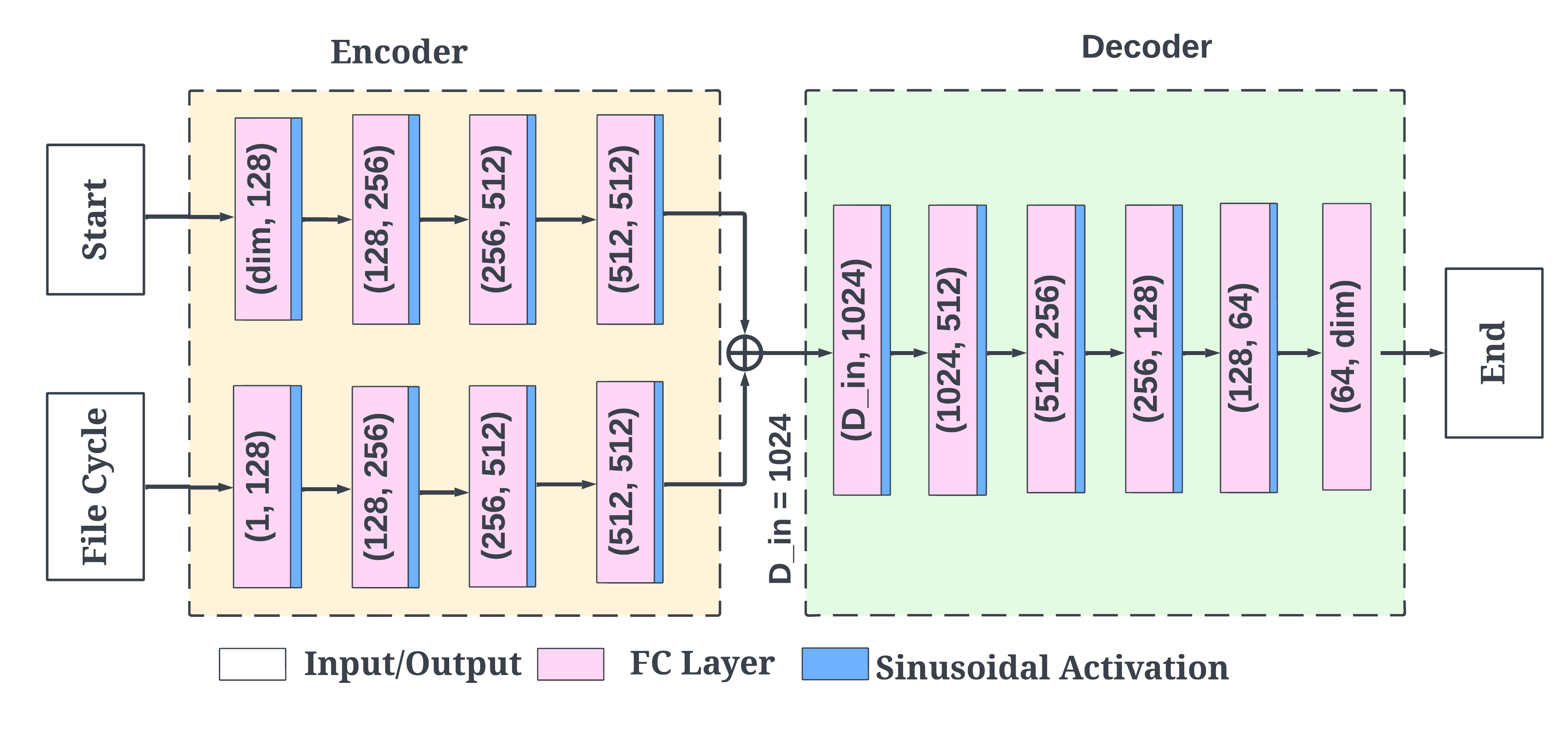}
    \caption{The MLP-based flow map NN proposed by Han et al.~\cite{han2024interactive}. This image, used with the authors' permission, illustrates a network configuration with four encoding layers, six decoding layers, and a latent vector dimension of 1024. The architecture begins by taking two inputs: the particle’s initial position (Start) and the number of file cycles (File Cycle). These inputs are first processed by the encoder, which transforms them into a latent vector denoted as $D\_in$. This latent vector is then passed to the decoder, which outputs the particle’s predicted position (End) at the queried file cycle. A sinusoidal activation function is applied after each fully connected (FC) layer, except for the output layer.}
    \label{fig:model}
\end{figure}

\subsubsection{Training Data Generation}
% To generate training data, Han et al.~\cite{han2021exploratory} introduced two basic tracing methods: Lagrangian\_long and Lagrangian\_short, and a Lagrangian\_hybrid~\cite{han2024interactive} that is a combination of long and short methods. 
%
In our work, we adopt only the Lagrangian\_long method introduced by Han et al.~\cite{han2024interactive}, which generates a single flow map by tracing long particle trajectories with uniform temporal sampling along each integral curve.
For seeding, we use a Sobol quasirandom sequence, which, as shown in previous work~\cite{han2021exploratory}, outperforms pseudo-random sequences and uniform grid sampling. 
Once the initial seeds are placed in the spatial domain, the particle trajectories are calculated by advancing them from time $t$ to $t + \delta$, where $\delta$ represents one simulation time step (or cycle). 
Tracing starts from the initial time $t0$ to the final time $T$, with the results saved at each file cycle. 
The final training dataset is structured as an $m \times n$ array, where $m$ is the number of seeds and $n$ is the number of file cycles. Each training sample contains a start position $s_i$ (where $0 \leq i \leq m -1$), the corresponding file cycle index $c_j$ (where $0 \leq j \leq n -1 $), and the end location ($\ell_{i,j}$).
The training dataset is formatted as \cref{eqn:input}, allowing the model to learn both spatial and temporal patterns in the flow field.

\begin{equation}
    \label{eqn:input}
    \begin{aligned}
    Input = & \{\{s_0, \,c_0, \,\ell_{0, 0}\}, 
              \{s_0, \,c_1, \,\ell_{0, 1}\}, ..., \\
             & \{s_0, \,c_{n-1}, \,\ell_{0, {n-1}}\}, ..., 
              \{s_{m-1}, \,c_{n-1}, \,\ell_{m-1, {n-1}}\}\}. 
    \end{aligned}
\end{equation}

\subsubsection{Network Architecture}
We adopt the MLP-based NN architecture proposed by Han et al.~\cite{han2024interactive} (see \cref{fig:model}).
The encoder \textbf{E} takes as input the particle start locations and the corresponding file cycles, processing them through two distinct sequences of fully connected (FC) layers.
The outputs of these sequences are concatenated to form a latent vector, which is then passed to the decoder \textbf{D}.
The decoder predicts the particle end locations, which are compared to the ground truth using the L1 loss.
The model uses the sine activation function throughout.
The network architecture is dynamic, featuring a configurable number of encoder and decoder layers, as well as a variable latent vector dimension.
%
% \jixian{To incorporate the MC Dropout method for uncertainty quantification, we append dropout after 1) each activation layer, 2) the last activation layer for all experiments using the MC Dropout method. We present a detailed construction and parameter study in \autoref{sec:hyperparameter_section}.}

\subsection{Uncertainty Quantification for Deep Neural Networks}
We employ three uncertainty quantification methods, Deep Ensembles, MC Dropout, and SWAG, to assess the uncertainty in the flow map prediction.

\subsubsection{Deep Ensembles Method}
\change{A deep ensemble~\cite{lakshminarayanan2017simple} consists of multiple, independently trained neural networks on the same dataset, each with different random initializations. This process leads each network to learn slightly varied representations of the data. During inference, an input is passed through every network, and the individual predictions are aggregated. The final prediction is the mean of these outputs, while the spread or disagreement among the predictions serves as a direct measure of the model's uncertainty. A high variance indicates low model confidence, whereas low variance suggests high confidence and model agreement.}

% A deep ensemble~\cite{lakshminarayanan2017simple} involves training multiple, independent NNs on the same dataset with different random initializations of network weights and varied data shuffling orders during the training process. This leads each network within the ensemble to learn slightly different representations and decision boundaries. Combined samples from these representations provide a more comprehensive understanding of the model's confidence in its predictions.

% After each NN has been trained, an input is passed through every member of the ensemble. Each network generates its prediction for the input. These individual outputs are then aggregated to form the final prediction and its uncertainty. The spread or disagreement among these individual predictions serves as a direct measure of the model's uncertainty. A high variance among the ensemble members' predictions indicates that the models are less confident, often because the input is in a region not well-represented in the training data. Conversely, a low variance suggests high model confidence and agreement.

\subsubsection{Monte Carlo Dropout}
\change{Monte Carlo Dropout (MC Dropout)~\cite{gal16dropout} estimates the uncertainty in NNs by keeping dropout active during inference. For each forward pass a random set of neuron is deactivated, enabling the single network to behave like an ensemble of multiple sub-networks without the computational burden of training multiple full models.  The MC Dropout predictive uncertainty is calculated by running multiple forward passes (Monte Carlo samples) for the same input and calculating the variation among the predictions.}

% Monte Carlo Dropout (MC Dropout)~\cite{gal16dropout} extends the dropout regularization techniques for quantifying uncertainty in NN. The dropout technique prevents overfitting by randomly deactivating a fraction of neurons during training.  MC Dropout extends this idea by keeping dropout active during the inference phase. This means that for a single input, the network's internal structure changes slightly with each forward pass due to the random deactivation of neurons, enabling the single network to behave like an ensemble of multiple sub-networks. The ensemble consequently provides an estimate of predictive uncertainty without the computational burden of training multiple full models.

% After training the NN with the dropout layers, the MC Dropout predictive uncertainty is calculated by running multiple forward passes (Monte Carlo samples) for the same input, with the dropout enabled for each pass. The variation of these predictions can then be used to represent the model's uncertainty.

\subsubsection{Stochastic Weight Averaging-Gaussian}
The Stochastic Weight Averaging-Gaussian (SWAG)~\cite{maddox_2019_swag} method is an innovative approach designed to quantify uncertainty in NNs while mitigating the high computational cost associated with traditional ensemble techniques. Unlike Deep Ensembles, which train multiple models independently, SWAG focuses on capturing a distribution over the NN's weights in a single training run.
It achieves this by averaging the network's weights throughout training, particularly towards the end of the optimization process, and then fitting a multivariate Gaussian distribution to these weights. The Gaussian approximates the Bayesian posterior of the model weights. An ensemble of models can be sampled from the Gaussian without requiring the training of multiple models.

To leverage the SWAG method for uncertainty quantification, an NN is first trained as usual. After the initial training phase, the learning rate is often set to a high constant value, and the network continues training using stochastic gradient descent (SGD) for a few more epochs to explore the loss landscape more comprehensively, thereby avoiding being stuck in local minima. During this fine-tuning phase, snapshots of the network's weights are periodically saved and averaged to ensure stability and consistency. Once this process is complete, a covariance matrix is estimated from these collected weight samples, often using a low-rank approximation to ensure computational feasibility. For a new input, predictions are then generated by sampling multiple sets of weights from this approximated Gaussian distribution, effectively creating a ``virtual ensemble" from a single trained model. The spread of these predictions, similar to a traditional ensemble, then indicates the model's predictive uncertainty.

% SWAG offers significant advantages by providing robust uncertainty estimates with a considerably lower computational overhead than full Deep Ensembles, as it avoids training multiple separate models from scratch. This makes it a more scalable solution for large-scale applications. Furthermore, the ability to approximate a Bayesian posterior allows for a more principled approach to uncertainty. However, SWAG also has limitations. Its performance can be sensitive to hyperparameter choices, such as the learning rate schedule and the frequency of weight averaging. While more efficient than Deep Ensembles, the process of collecting weight samples and estimating the covariance still adds complexity compared to a standard single-model training pipeline, and the quality of the uncertainty estimates depends on the accuracy of the Gaussian approximation to the true posterior.

\section{Visualizing Flow Map Uncertainty}
\begin{figure}[ht]
    \centering
    \begin{subfigure}[t]{0.5\columnwidth}
        \centering
        \includegraphics[width=1\columnwidth]{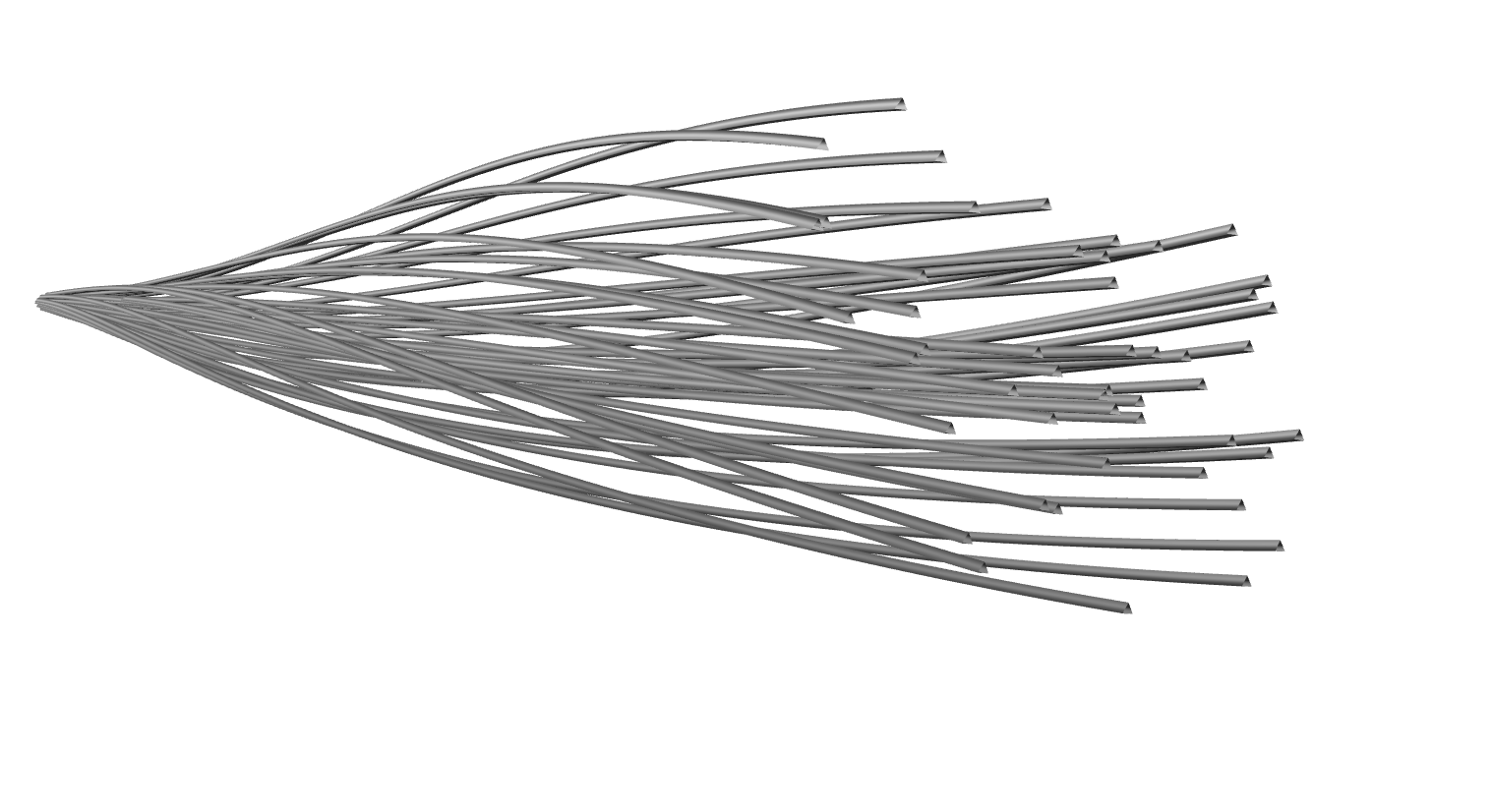}
        \caption{Spaghetti plot}
        \label{fig:demo_spaghetti}
    \end{subfigure}%
    \begin{subfigure}[t]{0.5\columnwidth}
        \centering
        \includegraphics[width=1\columnwidth]{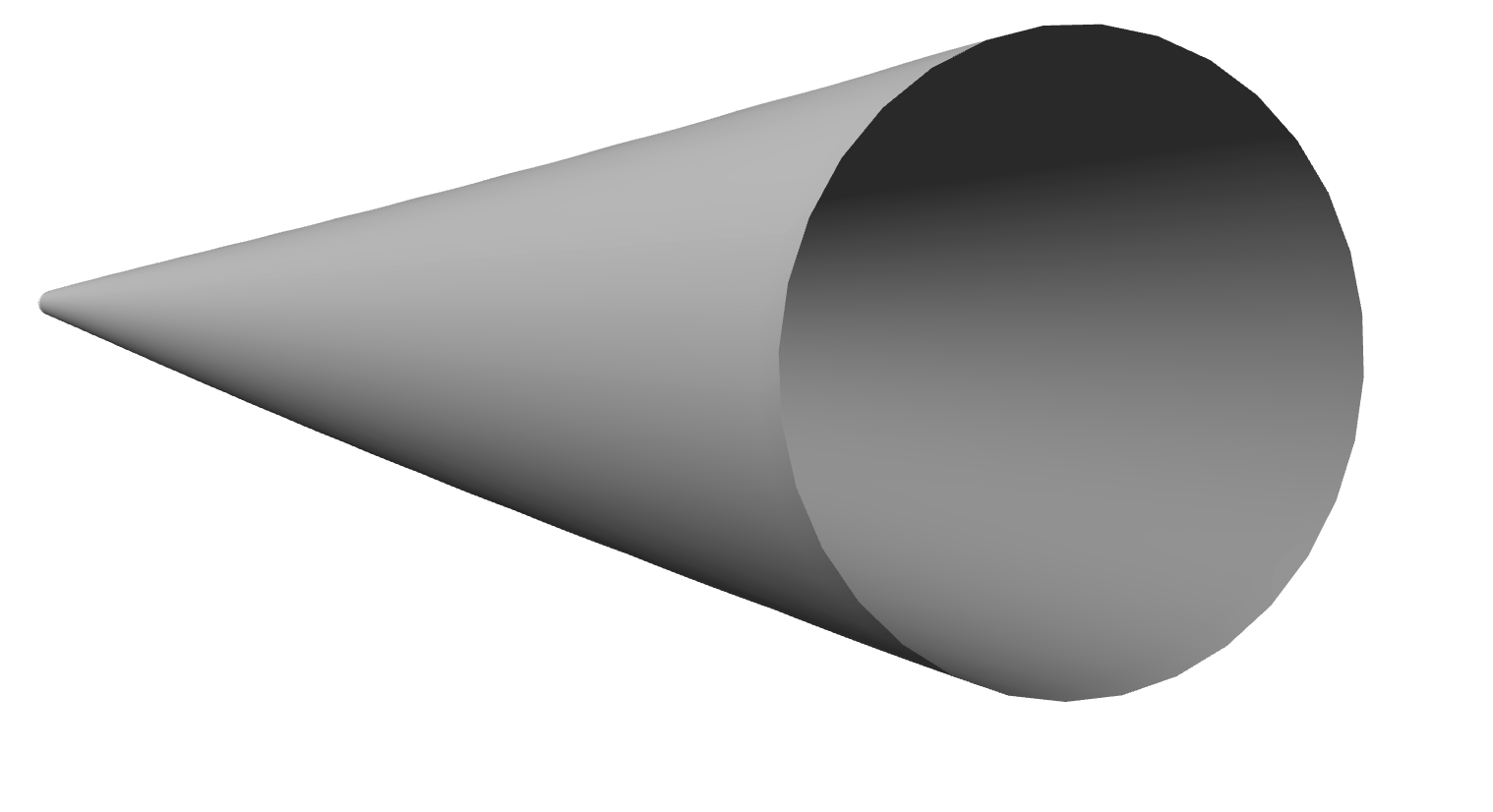}
        \caption{Circular tube}
        \label{fig:demo_circular_tube}
    \end{subfigure}
    \\
    \begin{subfigure}[t]{0.5\columnwidth}
        \centering
        \includegraphics[width=1\columnwidth]{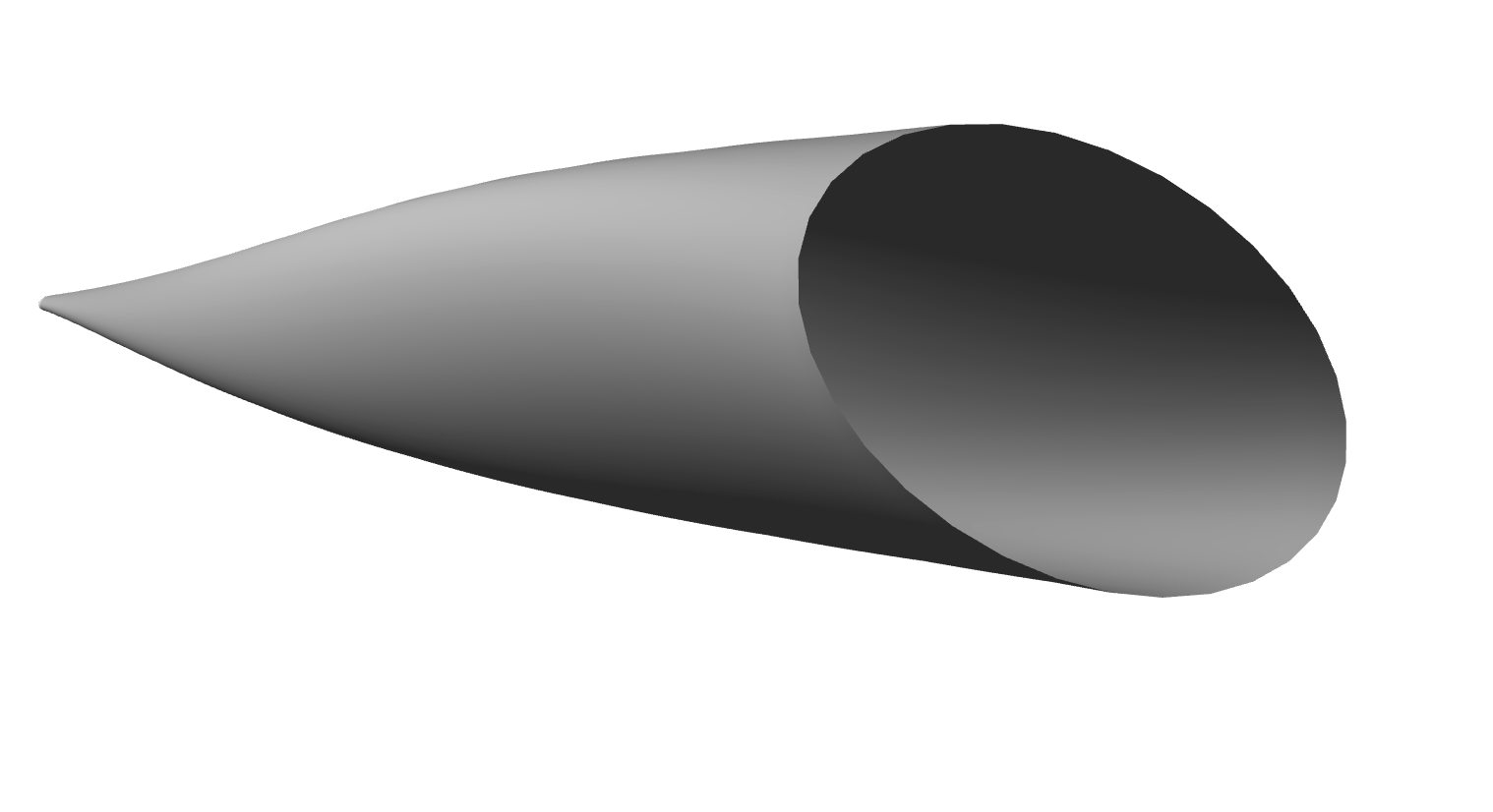}
        \caption{\textit{Uncertainty tube} with $\tau=2$}
        \label{fig:demo_ellipse}
    \end{subfigure}%
    \begin{subfigure}[t]{0.5\columnwidth}
        \centering
        \includegraphics[width=1\columnwidth]{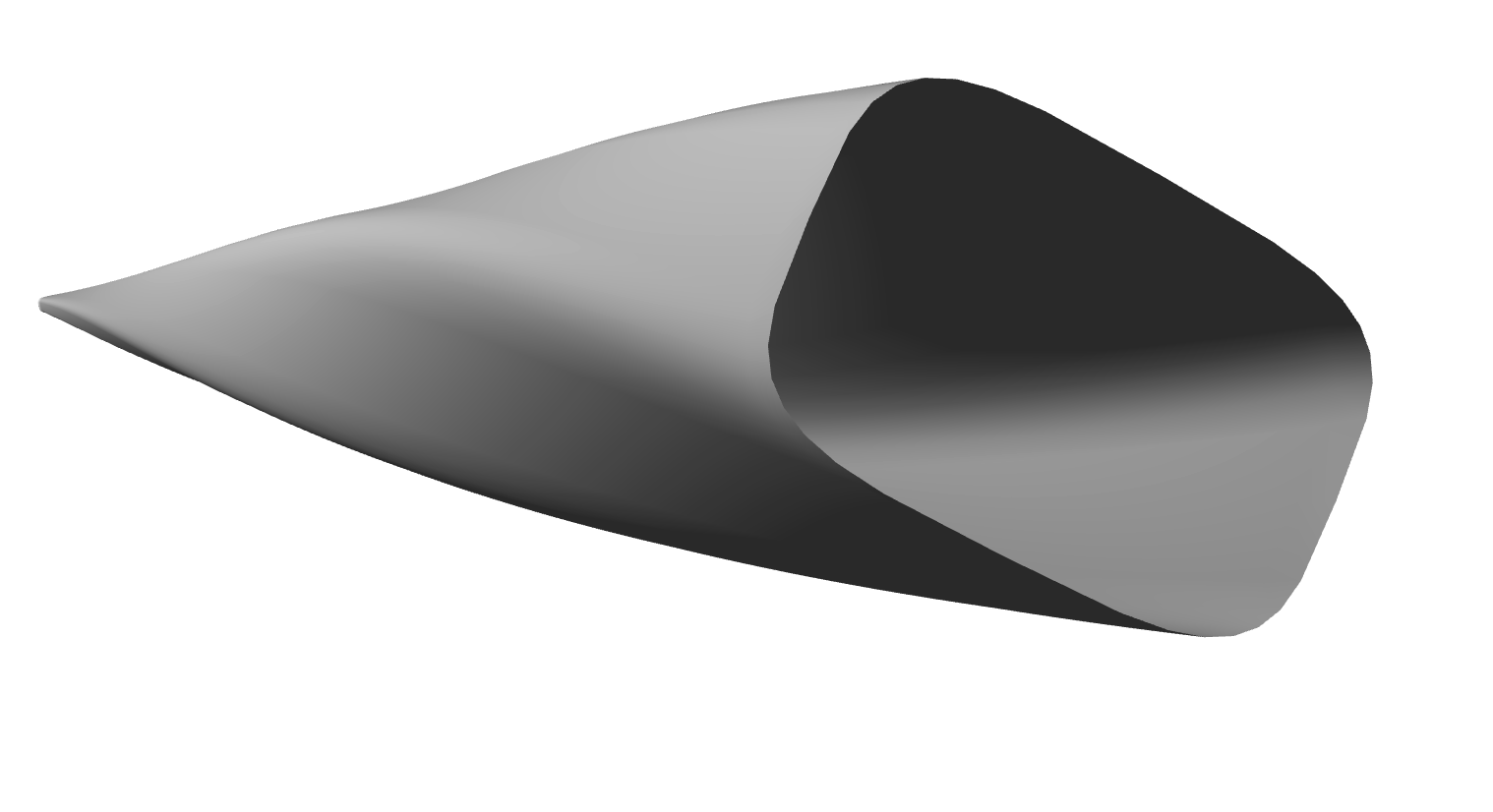}
        \caption{\textit{Uncertainty tube} with $\tau=4$}
        \label{fig:demo_superellipse}
    \end{subfigure}
    \caption{\change{For a given set of uncertainty samples (a), the circular tube (b) does not visually encode the twisty motion or asymmetric distribution. The elliptical tube (c) highlights the asymmetry with a small hint of twistiness. The superelliptical tube (d) highlights both asymmetry and twistiness.}}
    \label{fig:uq_overview}
\end{figure}

\subsection{Uncertainties of the Predicted Trajectories}
Kumar et al.~\cite{Kumar2025_vectorfield_uncertainty} visualize the integral line uncertainty using a circular tube, such that the variation is encoded as the radius of the tube. Upon closer examination of the uncertainty samples from multiple uncertainty quantification methods, we noticed that most of the uncertainty obtained from these methods is not symmetric. Visualizing them as a round tube hides the asymmetric nature of the uncertainty distribution. We observed the asymmetry of the uncertainty across models, datasets, and UQ methods. For some datasets, we were able to utilize the asymmetry to enhance our understanding of the data and improve the training of our model. More examples of how we can use this information are presented in \cref{sec:results}.

\change{\Cref{fig:demo_spaghetti} shows an example of nonsymmetrically distributed uncertainty. The gray uncertainty samples exhibit more uncertainty in one direction than the other. However, a spaghetti plot is not very effective here because of the complex occlusion patterns. Meanwhile, as \cref{fig:demo_circular_tube} demonstrates, a circular tube is not very effective in representing asymmetric and twisty uncertainty, either. We need a visual encoding that better captures the asymmetry and twistiness while reducing the visual clutter introduced by visualizing all uncertainty samples. We also need to compute the statistical summary sufficiently fast to avoid compromising the efficiency of our NN flow map model. Therefore, we introduce the \textit{uncertainty tube} as demonstrated in \cref{fig:demo_ellipse} and \cref{fig:demo_superellipse}.}
\begin{figure*}[!htb]
    \centering
    \begin{subfigure}[t]{0.50\columnwidth}
        \centering
        \includegraphics[width=\textwidth]{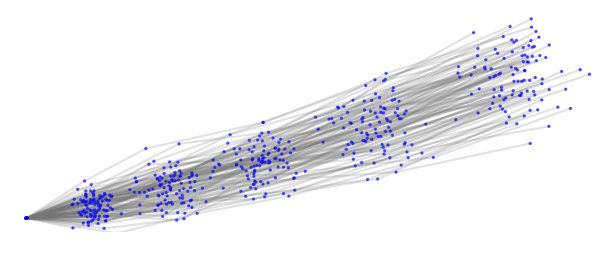}
        \caption{Step 1: Raw sample trajectories}
        \label{subfig:trajectories_and_points}
    \end{subfigure}
    \hfill
     \begin{subfigure}[t]{0.50\columnwidth}
        \centering
        \includegraphics[width=\textwidth]{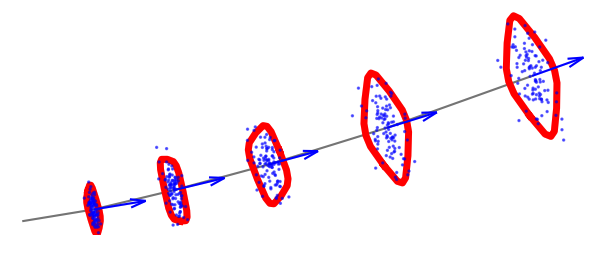}
        \caption{Step 2: Uncertainty superellipse}
        \label{subfig:uncertainty_super_ellipsoids}
    \end{subfigure}
    \hfill
     \begin{subfigure}[t]{0.50\columnwidth}
        \centering
        \includegraphics[width=\textwidth]{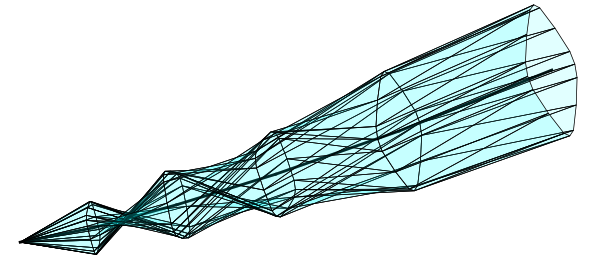}
        \caption{Step 3: Tube mesh before alignment}
        \label{subfig:unaligned_tube}
    \end{subfigure}
    \hfill
    \begin{subfigure}[t]{0.50\columnwidth}
        \centering
        \includegraphics[width=\textwidth]{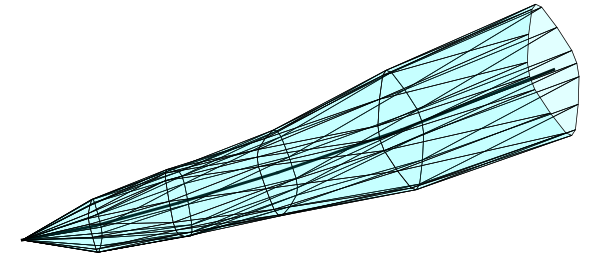}
        \caption{Step 4: \textit{Uncertainty tube} mesh}
        \label{subfig:uncertainty_tube}
    \end{subfigure}
    \caption{\change{\textit{Uncertainty tube} construction steps. This figure illustrates the design process for constructing the \textit{uncertainty tube}, starting from the raw trajectories shown in (a), followed by the projection to form superellipses in (b), and finally the alignment step in (c) and (d) to form the \textit{uncertainty tube}.}}
    \label{fig:uncertainty_tube_computation_demo}
    \vspace{-1em}
\end{figure*}

\subsection{Uncertainty Tube}
\label{subsec:uncertainty-tube}

\change{The \textit{uncertainty tubes} are constructed by building a superelliptical tube between two consecutive time steps, $t-\delta$ and $t$, starting from the seed location. For an ensemble of particle trajectories with $N$ time steps, as illustrated in \cref{subfig:trajectories_and_points}, we start at the seed location $\mathbf{x}^{(0)}$, which is assumed to have no uncertainty. For each subsequent time $t = \{{\delta, \cdots, N\delta}\}$, We build the super elliptical tube between $t-\delta$ and $t$. Let $\mathbf{x}_i^{(t)}$ be uncertainty samples at $t$ and $\mathbf{\bar{x}}^{(t)}$ be the mean of those points. We project all $\mathbf{x}_i^{(t)}$ to the plane orthogonal to the direction from $\mathbf{\bar{x}}^{(t-\delta)}$ to $\mathbf{\bar{x}}^{(t)}$ and passing through $\mathbf{\bar{x}}^{(t)}$. The projection is defined by}
\begin{equation}
    \mathbf{p}_{i}^{(t)} = \mathbf{x}_{i}^{(t)} - ((\mathbf{x}_{i}^{(t)} - \mathbf{\bar{x}}^{(t)})\cdot \mathbf{d})\mathbf{d}, 
\end{equation}
\change{where $\mathbf{d}$ is the unit normal vector to the plane. This projection is designed to capture the uncertainty in the orthogonal cross-section, but does not account for the variation along the normal direction ($\mathbf{d}$).}

We calculate the covariance matrix of the projected points and its eigenvalue decomposition:
\begin{equation}
    \mathbf{V} \Sigma \mathbf{V}^{T} = \frac{1}{N} \sum_{i=1}^{N}(\mathbf{p}_{i}-\mathbf{\bar{p}}) \cdot (\mathbf{p}_{i}-\mathbf{\bar{p}})^{T},
    \label{eq:eig_cov}
\end{equation}
where $\Sigma$ and $\mathbf{V}$ are the eigenvalues and vectors, respectively. 
The right-hand side of \cref{eq:eig_cov} shows the covariance calculation with $\bar{\mathbf{p}}$ being the mean of the projected points. The result from the decomposition is then used to construct a superellipse according to
\begin{equation}\label{eq:super-ellipse}
    \mathbf{q}(\theta) = \mathbf{e}(\theta)\mathbf{V}^{T} + \mathbf{\bar{p}},
\end{equation}
where the function $e(\theta)$ is defined according to
\begin{equation}
    \mathbf{e}(\theta)= 
    \begin{pmatrix}
        2\sigma_{1}|cos(\theta)|^{\frac{2}{\tau}}  sgn(cos(\theta)) \\
        2\sigma_{2}|sin(\theta)|^{\frac{2}{\tau}}  sgn(sin(\theta)) \\
    \end{pmatrix}, 
    \theta \in [0, 2\pi],
\end{equation}
where $\sigma_{1}$ and $\sigma_{2}$ are the diagonal entities of $\Sigma$. The function $sgn$ returns the sign of its input value. The parameter $\tau \geq 2$ controls the shape of the superellipse; For greater values of $\tau$ the superellipse becomes more rectangular with sharp corners, whereas for $\tau=2$ the shape reduces to the standard ellipse. The results in this paper use $\tau=4$. The superellipse centered at the mean $\mathbf{\bar{p}}$ summarises the variance of the projected points. The superellipse representation provides a more accurate representation of the direction variation of the projected points compared to a standard circle. Additionally, its rectangular shape helps clarify orientation more effectively than a standard ellipse. These uncertainty superellipses are shown in \cref{subfig:uncertainty_super_ellipsoids}. To form the \textit{uncertainty tube}, the superellipses at $t$ and $t + \delta$ are sampled, and their corresponding boundary points are connected. 
 
However, the sampled points $\mathbf{q}_{j}^{(t)}$ and $\mathbf{q}_{j}^{(t+ \delta)}$ can be misaligned, leading to warped and twisted superelliptical tubes, as shown in \cref{subfig:unaligned_tube}. To address this, we calculate an optimal circular shift and reverse orientation ordering that minimizes the alignment score according to

\begin{equation}\label{eq:rot-alignment}
\begin{split}
    \{\hat{r}, \hat{s}\} &=\operatorname*{argmin}_{r \in \{0,1\}, s \in \{0, 1, \cdots, m-1\}} \sum_{j =1}^{m}\|\mathbf{q}^{(t+\delta, r)}_{\ell_{j}} - \mathbf{q}^{(t)}_{j}\|, \\
    \ell_{j}&= (j+s)\textrm{mod } m.
\end{split}
\end{equation}
The subscript $\ell_{j}$ in \cref{eq:rot-alignment} performs the circular shift at $s$, and a reverse ordering is employed if $r=1$. The aligned \textit{uncertainty tube} is shown in \cref{subfig:uncertainty_tube}. 

\change{Overall, the \textit{uncertainty tube} significantly improves the representation of asymmetric uncertainty compared to the circular tube. Moreover, the superelliptical tube's rectangular design distinctly shows the uncertainty orientation and its evolution along the pathline more effectively than the circular and elliptical tubes, as demonstrated in \cref{fig:uq_overview}.}

\subsection{Computation Efficiency}
The computation of the \textit{uncertainty tube} introduces a one-time cost per user's query.
\change{The overhead mainly consists of two parts: 1. the \textit{UQ time} measures the time it takes to obtain the uncertainty samples. 2. The \textit{meshing time} is the time it takes to compute the mesh and texture coordinates.}

\change{The UQ time is reported in \cref{sec:controlled_experiment_section} for different methods. Here, we report the meshing time} on AMD Ryzen Threadripper 3970X 32-Core Processor using 32-core parallelization in \cref{tab:perf}. Each row represents the number of seeds. Each column represents the number of steps per trajectory. To compute the \textit{uncertainty tube}, we use 50 uncertainty samples per trajectory. The number of uncertainty samples within the range of 10 to 100 has a minimal impact on the meshing time. 

A typical setup in our data exploration stage uses 100 to 300 seeds to understand the global trend. All our datasets have 50 to 150 steps. We found that 30 to 50 uncertainty samples are sufficient to build a meaningful statistical model of the uncertainty. This means that \textit{uncertainty tube} computation incurs a few seconds of overhead per query, which still fits within the interactive exploration scheme proposed in Han et al.~\cite{han2024interactive} given ensemble samples from the UQ method. 

We normally spend a much longer time interacting with the rendered scene than sending different queries. In our experiments, all \textit{uncertainty tube} visualization renders at 120 frames per second on a 2023 MacBook Pro with an Apple M2 Max chip after the initial rendering pass. 
\begin{table}
\centering
\begin{tabular}{|c|ccccc|}
\hline
Seeds/Steps & 10 & 50 & 100 & 150 & 200 \\
\hline
10 & 401 & 440 & 561 & 608 & 653 \\
\hline
100 & 625 & 744 & 868 & 971 & 1124 \\
\hline
300 & 936 & 1220 & 1667 & 1823 & 2122 \\
\hline
500 & 1029 & 1509 & 1957 & 2454 & 2994 \\
\hline
\end{tabular}
% \begin{tabular}{|c||ccccc|}
% \hline
% Seeds/Steps & 10 & 50 & 100 & 150 & 200 \\
% \hline
% \hline
% 10 & 259 & 306 & 456 & 409 & 506 \\
% \hline
% 100 & 540 & 645 & 730 & 883 & 987 \\
% \hline
% 300 & 798 & 1033 & 1576 & 1703 & 2043 \\
% \hline
% 500 & 1087 & 1475 & 1959 & 2425 & 3082 \\
% \hline
% \end{tabular}
\caption{\change{\textit{Uncertainty tube} computation costs in milliseconds (ms)}}
\label{tab:perf}
\end{table}

\subsection{Uncertainty Coloring}
\change{Sometimes the geometry alone is not sufficient for the analysis of uncertainty. For example, comparing \cref{fig:halfcylinder} and \cref{fig:halfcylinder_rescaled} without color is challenging because the camera is positioned far away to reveal the dataset's global pattern, and the size difference is visually diminished. Therefore, we use color to help us compare uncertainties at a different granularity.}

\begin{figure*}[ht]
    \begin{subfigure}[t]{0.33\linewidth}
        \centering
        \includegraphics[width=1\textwidth]{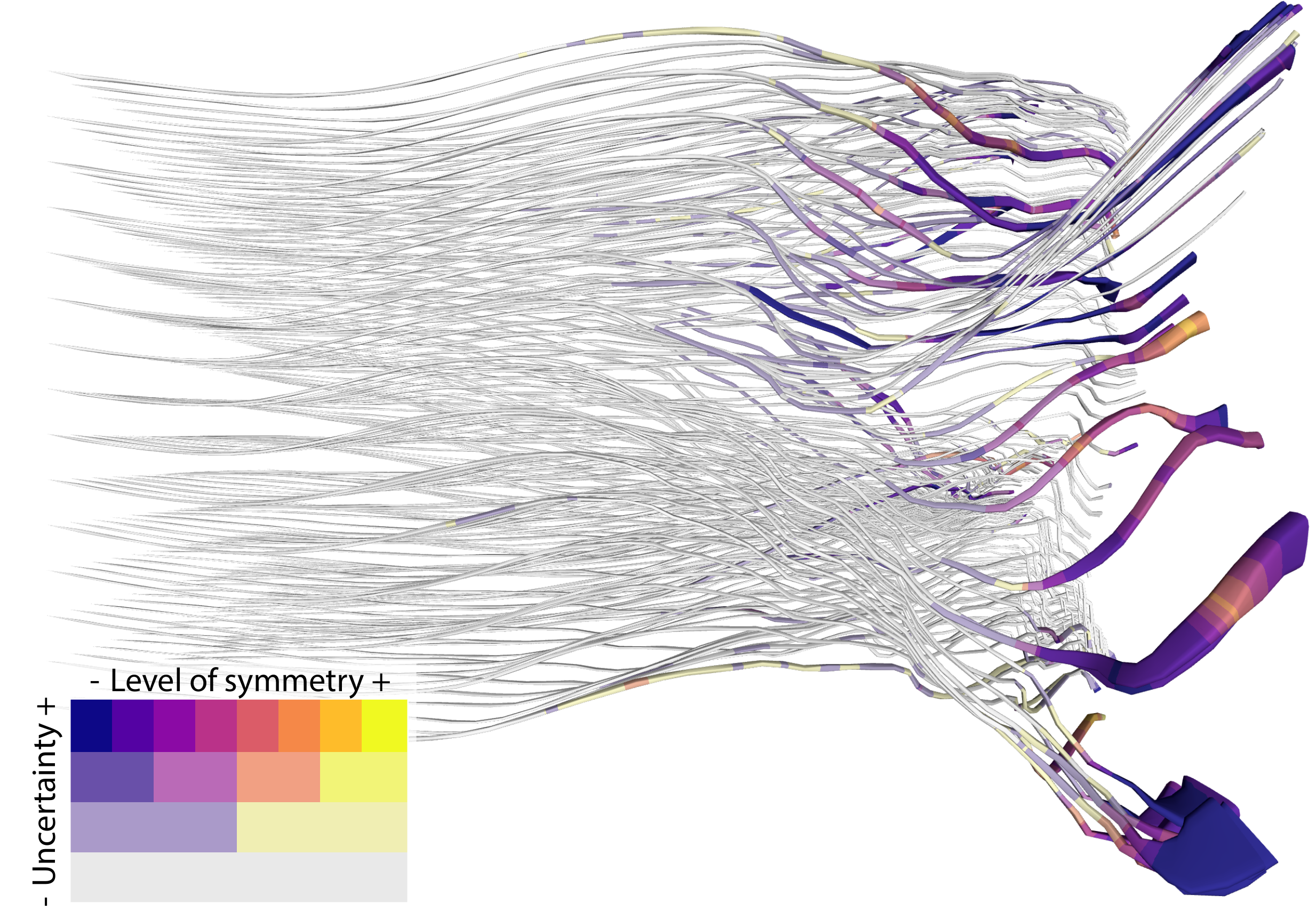}
        \caption{Deep Ensembles}
        \label{fig:synth_de}
    \end{subfigure}
    \hfill
    \begin{subfigure}[t]{0.33\linewidth}
        \centering
        \includegraphics[width=1\textwidth]{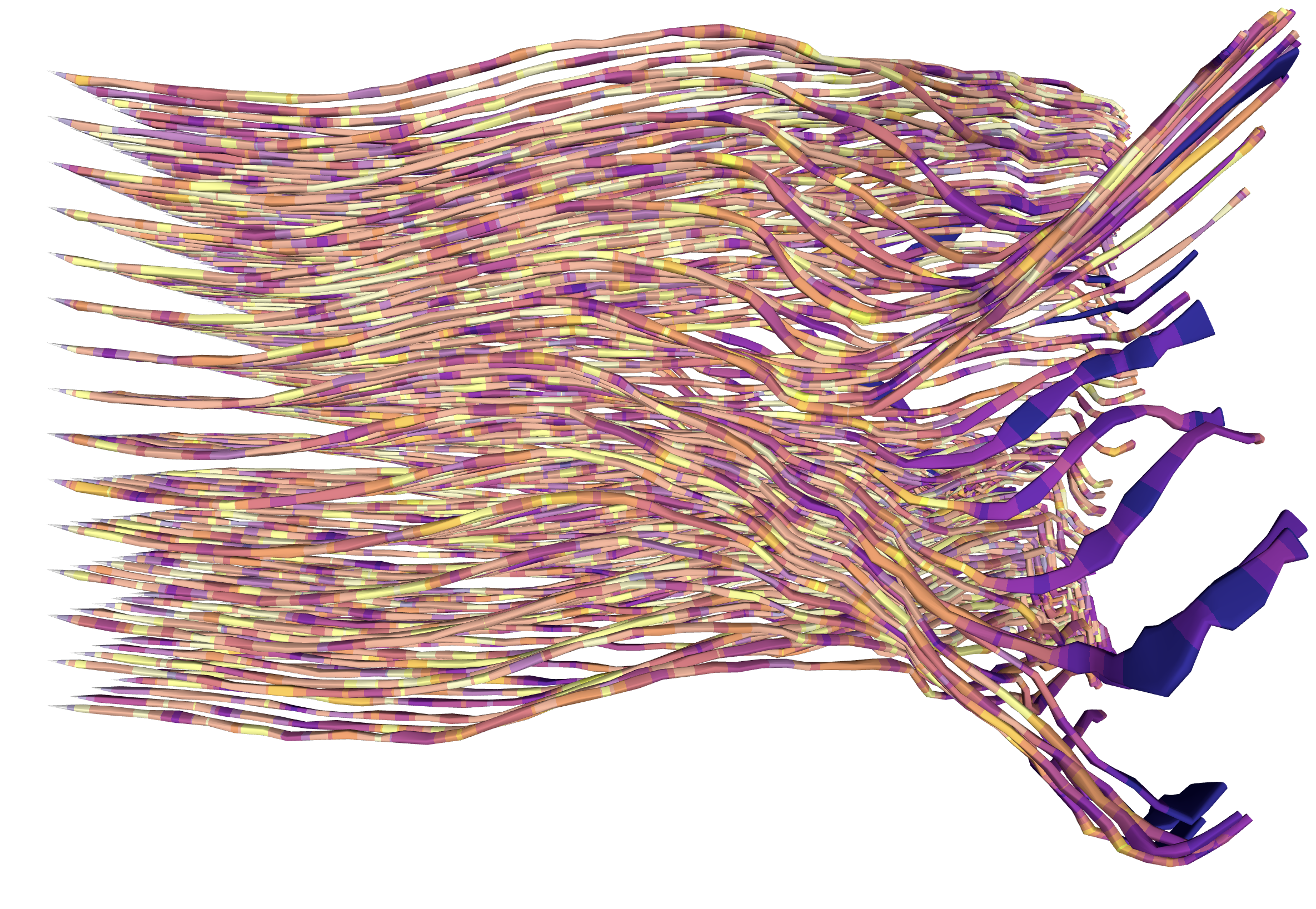}
        \caption{MC Dropout}
        \label{fig:synth_mc}
    \end{subfigure}%
    \hfill
    \begin{subfigure}[t]{0.33\linewidth}
        \centering
        \includegraphics[width=1\textwidth]{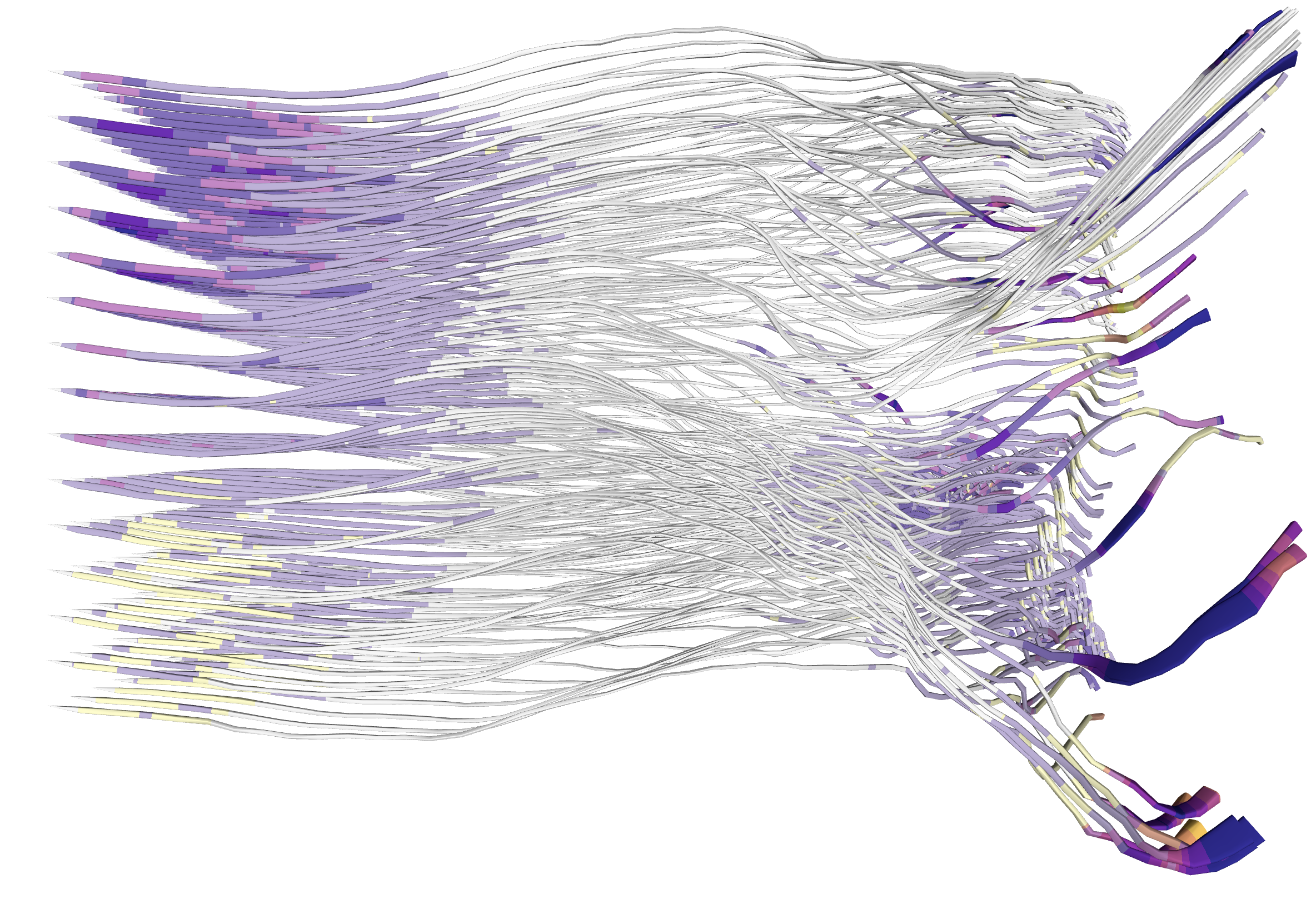}
        \caption{SWAG}
        \label{fig:synth_swag}
    \end{subfigure}
    \caption{\textit{Uncertainty tube} visualizations of 225 pathlines from \synth~dataset using three quantification methods. The $z$ value, hence the introduced uncertainty, increases from left to right. Each tube is computed from 50 uncertainty samples.}
    \label{fig:synth_ut}
\end{figure*}

Inspired by value-suppressing uncertainty palettes (VSUP)~\cite{correl2018vsup}, we employ a similar color map to represent the amount of uncertainty and the level of symmetry. Our primary task is to visualize the uncertainty of an ensemble of trajectories. 
Unlike the original VSUP, we suppress trajectories with low uncertainty; that is, if the prediction has low uncertainty, the colormap does not distinguish between the levels of symmetry. 
We chose a light gray color, indicating a low level of uncertainty. For high uncertainty, we use a linear color palette to determine the level of symmetry. 
The level of symmetry is determined by the ratio of the first two eigenvalues. $1$ means symmetry, where the \textit{uncertainty tube} would be round, and $0$ means high asymmetry, indicating one major variation direction, resulting in a flat \textit{uncertainty tube}. 
The uncertainty colormap first determines the color for the level of symmetry by linearly interpolating the color palette. Then the final color can be computed by interpolating between the color of symmetry and light gray according to the level of uncertainty. 
The level of uncertainty is the magnitude of the first eigenvalue, rescaled to a value between 0 and 1, according to the user-set threshold. 
For example, the user could map the 98th percentile of the data to 1 to reduce the impact of extreme values. In the visualization, using the viridis colormap as an example, gray represents low uncertainty, blue represents nonsymmetric uncertainty, and yellow represents symmetric uncertainty. \Cref{fig:synth_ut} demonstrates how our colormap is applied in a synthetic dataset.

\section{Controlled Experiments}
\label{sec:controlled_experiment_section}
\begin{figure}
    \begin{subfigure}[t]{0.48\columnwidth}
        \centering
        \includegraphics[width=\columnwidth]{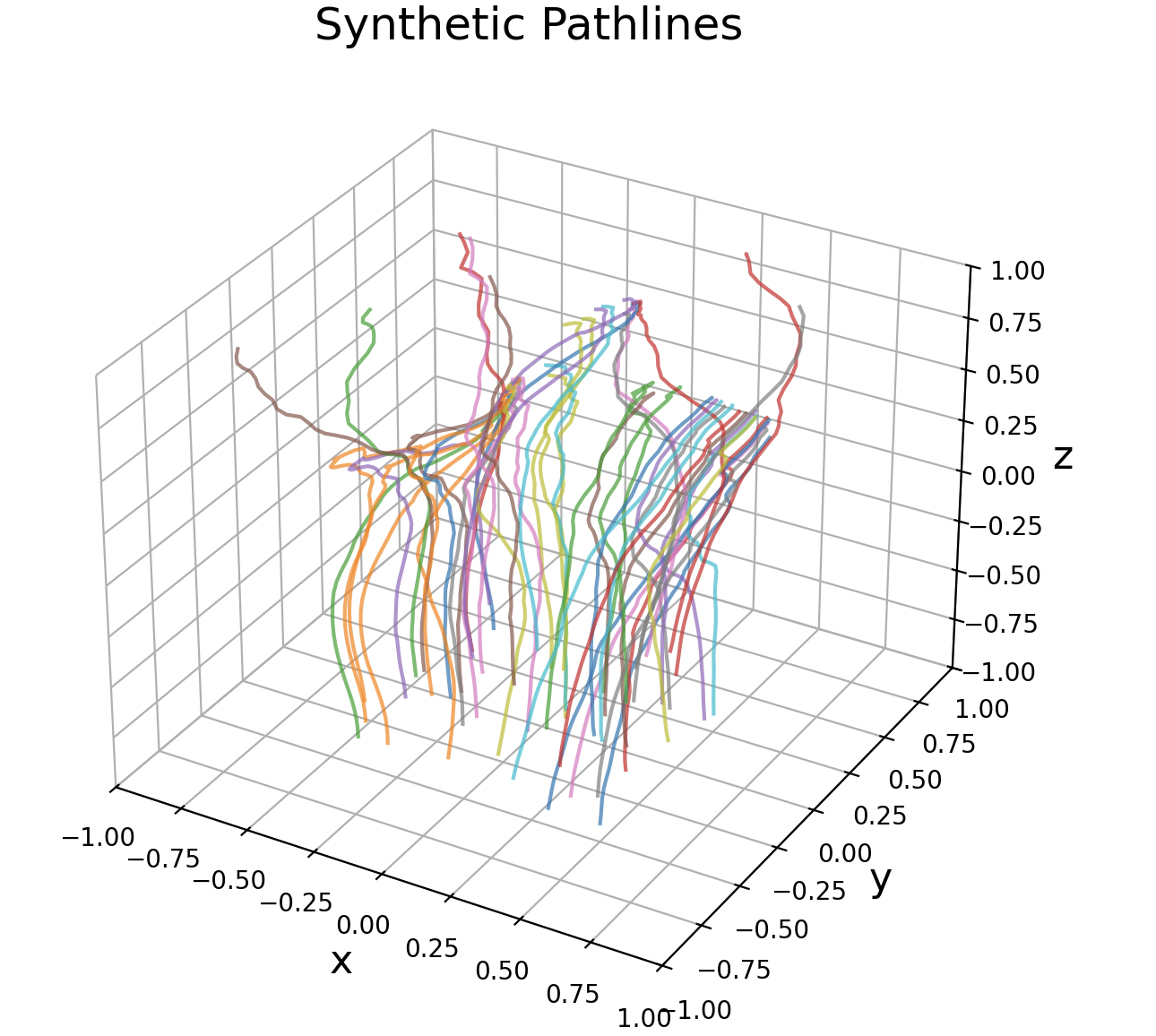}
        \caption{50 random pathlines}
        \label{fig:synth_dataset}
    \end{subfigure}
    \hfill
    \begin{subfigure}[t]{0.48\columnwidth}
        \centering
        \includegraphics[width=\columnwidth]{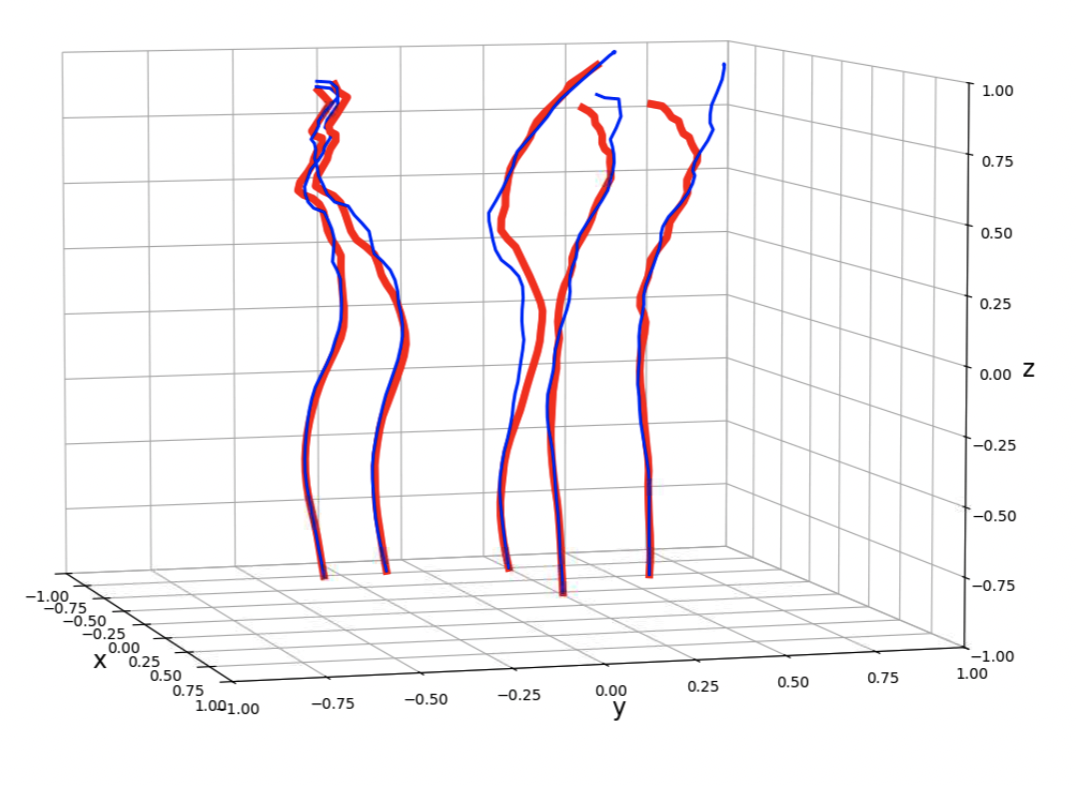}
        \caption{Test data (red) compared to model prediction(blue)}
        \label{fig:synth_test_demo}
    \end{subfigure}
    \caption{Matplotlib demonstration of the \synth~dataset and training outcomes.}
\end{figure}
\begin{figure}[htb]
    \begin{subfigure}[t]{0.3\columnwidth}
        \centering
        \includegraphics[width=1\textwidth]{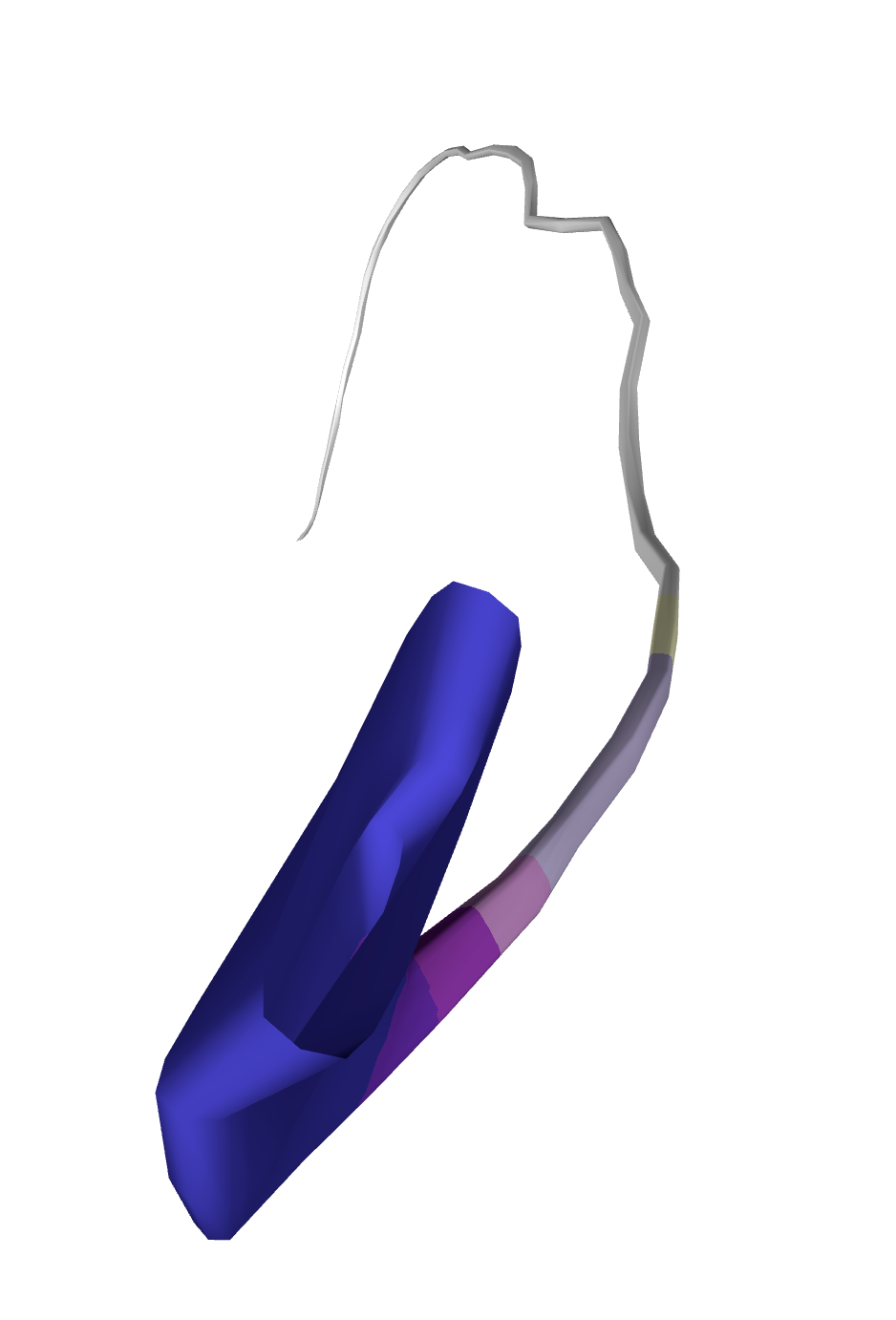}
        \caption{Deep Ensembles}
        \label{fig:synth_one_de}
    \end{subfigure}
    \hfill
    \begin{subfigure}[t]{0.3\columnwidth}
        \centering
        \includegraphics[width=1\textwidth]{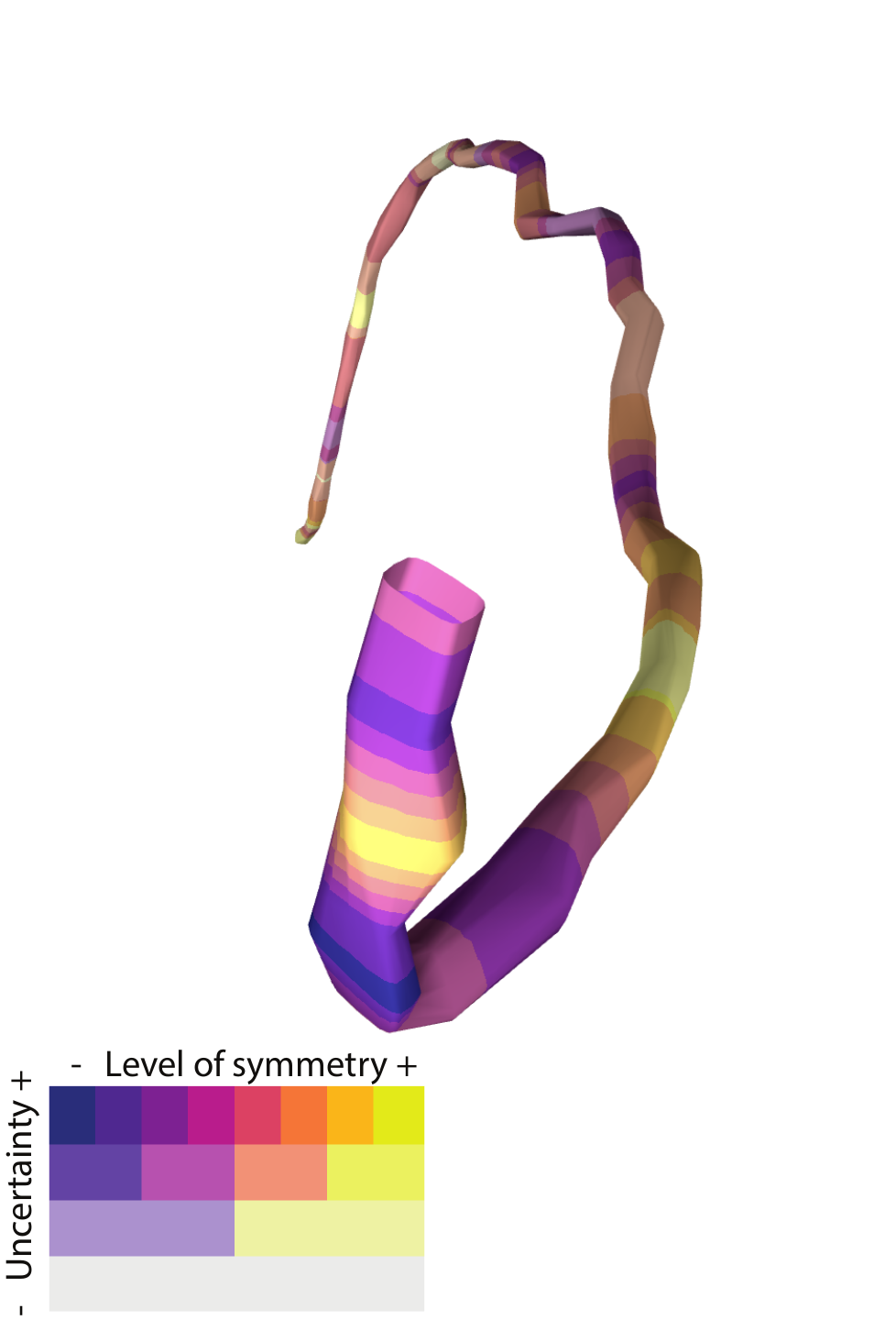}
        \caption{MC Dropout}
        \label{fig:synth_one_mc}
    \end{subfigure}%
    \hfill
    \begin{subfigure}[t]{0.3\columnwidth}
        \centering
        \includegraphics[width=1\textwidth]{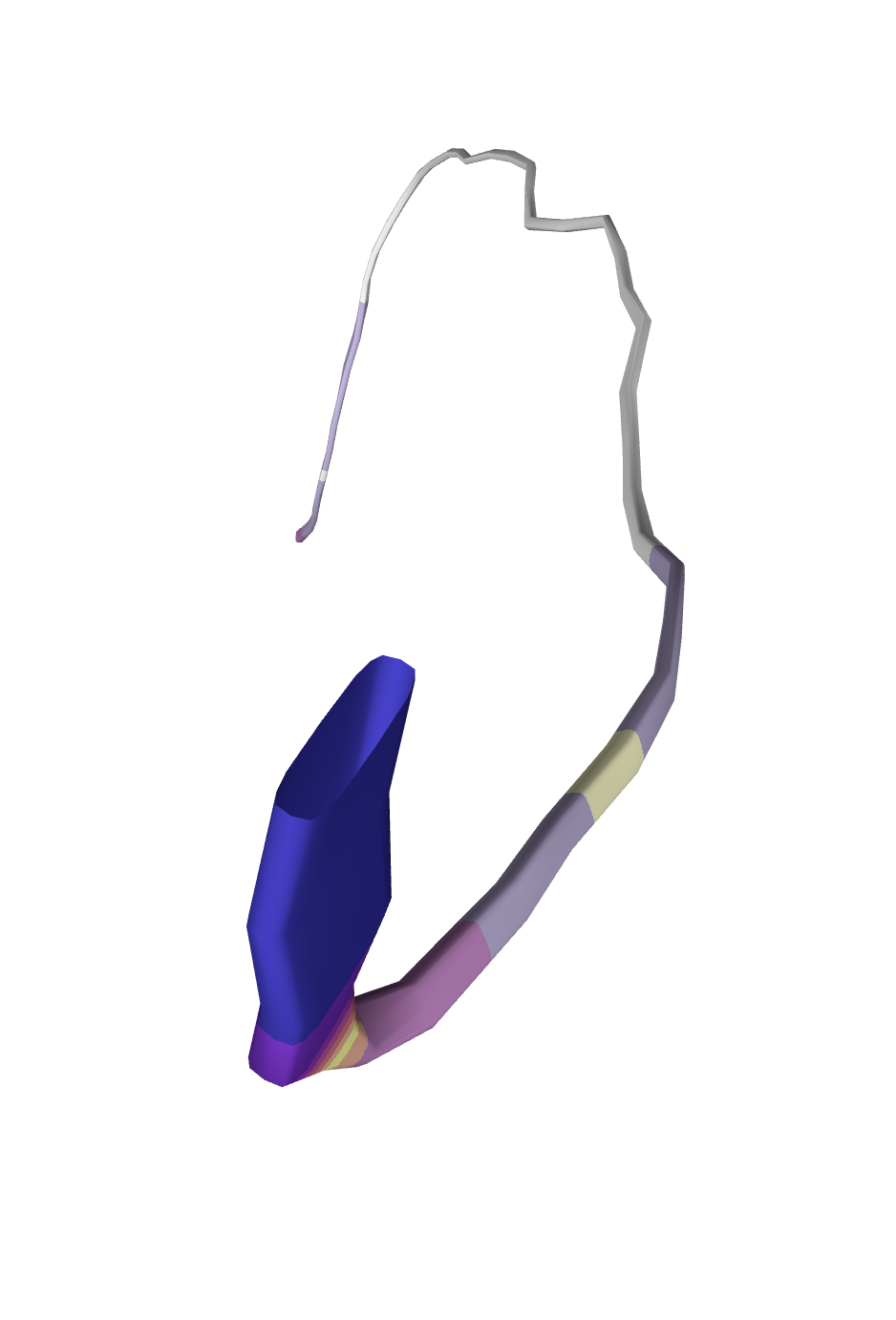}
        \caption{SWAG}
        \label{fig:synth_one_swag}
    \end{subfigure}
    \caption{\textit{Uncertainty tube} visualizations of one pathline from \synth~dataset that has a large variation at the end using three quantification methods. The seed location is at (-0.45,-0.1,-0.95).}
    \label{fig:synth_one_ut}
\end{figure}

We first use a synthetic dataset, labeled as \synth, to demonstrate different uncertainty quantification and visualization in a controlled setting. 
The \synth~dataset is a time-varying vector field in which the particles move in the positive $z$-direction. We introduce more complexity as the $z$ value increases. 
Hence, we expect the trained model to exhibit more uncertainty in the positive $z$-direction. 
The domain is $[-1, 1]$ for all axes.
The seeding box is $[-0.5, 0.5], [-0.5, 0.5]$, and $[-1, -0.9]$ for the x, y, and z axes, respectively. Each pathline is traced 50 steps, including the seeds. \cref{fig:synth_dataset} shows 50 random pathlines traced from the seeding box. 
For the training data, we use the Sobol method to generate $131,072$ seed locations. The testing dataset consists of $5,000$ uniformly sampled seeds inside the seeding box.

\subsection{Deep Ensembles}

To utilize the Deep Ensembles method, we independently trained $50$ models on the same training dataset. To induce randomness in the training process, we randomly shuffle the order of the training dataset at each iteration. For each model, we use four encoder layers and four decoder layers, setting the latent dimension to $1024$. We train a total of $10,000$ iterations. On two NVIDIA 3090 GPUs, training each model takes approximately $3.5$ minutes. The total time for training $50$ models is approximately 3 hours. Quantifying uncertainty using the Deep Ensembles has a significant computational time overhead. However, we also notice that the Deep Ensembles method reports uncertainty most faithful to our construct. We did not try to optimize the training parameters for the optimal training quality. On average, we expect an absolute difference of $0.027$ between the predicted location and the truth. \Cref{fig:synth_test_demo} shows the quality of the trained model on five randomly selected testing pathlines. As expected, the error increases as $z$ increases.

In practice, we often lack the testing data to assess the true error of the predictions. Thus, we rely on the uncertainty quantification to provide insight into the model's confidence. We sample $225$ points within the seeding box and then evaluate the pathlines from all $50$ models.
We calculate the mean pathline across 50 models and use the $51$ paths per seed to build the \textit{uncertainty tube}. \Cref{fig:synth_de} shows the \textit{uncertainty tube} visualization, from which we observe that the models exhibit higher variation in the positive $z$ direction (rightward), aligning with our expectation. We also noticed that in the high-uncertainty regions, the distribution of uncertainty trajectories is mostly asymmetric, as indicated by the predominance of blue hues toward the end. We selected one pathline with high uncertainty from the scene to demonstrate the bias in \cref{fig:synth_one_de}.

In summary, the Deep Ensembles method provides a good demonstration of uncertainty, aligning with our data constructions, but at a relatively high computational cost, as noticed in other works that utilize the Deep Ensembles method.

\subsection{MC Dropout}
% \begin{figure}
%     \centering
%     \includegraphics[width=\columnwidth]{figures/synth_mc.png}
%     \caption{The uncertainty tube visualization of MC Dropout for the \synth~dataset at dropout rate 0.02 for all layers.}
%     \label{fig:synth_mc}
% \end{figure}
\begin{table*}[htb]
    \centering
    \begin{tabular}{|c|c|c|c|c|c|c|c|c|}
    \hline
        dropout rates   &  0.1  & 0.05  & 0.04  & 0.03  & 0.02  & 0.01  & 0.005 & 0.001\\\hline
        all layers      &  0.034& 0.031 & 0.030 & 0.029 & 0.029 & 0.028 & 0.028 & 0.027 \\
        last layer      &  0.032& 0.028 & 0.029 & 0.029 & 0.028 & 0.028 & 0.028 & 0.027\\
        \hline
    \end{tabular}
    \caption{Test errors (absolute differences) for \synth~ of different dropout methods and rates.}
    \label{tab:mc_dropout_error}
\end{table*}
MC Dropout introduces the least amount of overhead among the three UQ methods used in this paper, provided that dropout is already part of the model during training. Evaluating 225 trajectories with 50 uncertainty samples takes less than 400 milliseconds.
It requires modifying the model by appending dropout layers if dropout is not already part of the training. 

To conduct the MC Dropout experiments, we implemented two ways of appending dropout layers. 
The first way is to append a dropout layer after every activation. The second approach is to append a dropout layer after the last activation. 
\Cref{tab:mc_dropout_error} demonstrates the impact of adding dropout on the testing metric, namely, the absolute difference between the prediction and the truth. 
We confirm the finding in Kumar et al.~\cite{Kumar2025_vectorfield_uncertainty} that adding dropout results in a decrease in prediction quality. For our model, dropping out after all activation layers at a rate of 0.001 has minimal impact on prediction quality, and it closely matches the original proposed MC Dropout method \cite{gal16dropout}. We use this configuration to produce the outcomes shown in \cref{fig:synth_mc}.

\Cref{fig:synth_mc} shows the pathlines evaluated from the same set of seeds used to demonstrate the Deep Ensembles method. \change{The magnitude of uncertainty estimated by MC Dropout is larger than Deep Ensembles' result in the low-uncertainty region, as shown by the colorfulness of the \textit{uncertainty tubes}. At the same time, the high-uncertainty region (bottom right) shows a smaller uncertainty magnitude compared to the result from Deep Ensembles. However, overall, the MC Dropout exhibits higher uncertainty at larger z values, as demonstrated by the change in size of the \textit{uncertainty tubes}. \cref{fig:synth_one_mc} also shows that MC Dropout's result may be significantly different than the results of the other two methods. } 

In summary, MC Dropout is a straightforward method that incurs no additional training costs. Due to neural networks' efficient inference capability, the overhead of running multiple inference passes is minimal. However, while originally proposed as an approximate Bayesian method, it is difficult to rigorously argue that the samples from MC Dropout are drawn from a true Bayesian posterior distribution. This is because dropout randomly sets activations to zero, and the underlying theoretical approximations are often not strictly met in practice. Indeed, research by Folgoc et al. \cite{folgoc2021mc} has strongly argued that MC Dropout does not perform approximate Bayesian inference.

\subsection{SWAG}
Maddox et al. \cite{maddox_2019_swag} introduced the SWAG method in 2019. Despite its simple and efficient utilization, we have yet to see it being used for visualizing uncertainty in the visualization community. 
This method introduces minimal overhead and does not require modification to the model. However, it requires careful tuning of hyperparameters. We report a hyperparameter study in \cref{sec:hyperparameter}.

We generated \cref{fig:synth_swag} to demonstrate the uncertainty estimated by the SWAG method. SWAG requires some additional training from a pre-trained model. Training 1000 steps using SGD takes 15 seconds on a NVIDIA 3090. Thus, we can quickly experiment with different hyperparameters in our applications.

\Cref{fig:synth_swag} shows that our SWAG model reports slightly higher uncertainty at the beginning (left) and lower uncertainty at the end compared to Deep Ensembles. However, the overall trend aligns with the result of Deep Ensembles. Especially, \cref{fig:synth_one_swag} shows that the uncertainty direction matches closer to the Deep Ensembles result, compared to the MC Dropout result.

In summary, SWAG gives a descent uncertainty quantification of the model, and it is easy to set up and use. Although it requires careful hyperparameter tuning, this process is not overly complicated. We recommend that users start with a rank of 100 and find the ideal \swaglr~and \nswagsamples~first, then adjust other parameters as needed. 

\subsubsection{SWAG hyperparameter study}
\label{sec:hyperparameter}
\begin{figure*}[htb]
    \begin{subfigure}{0.3\textwidth}
        \centering
        \includegraphics[width=1\textwidth]{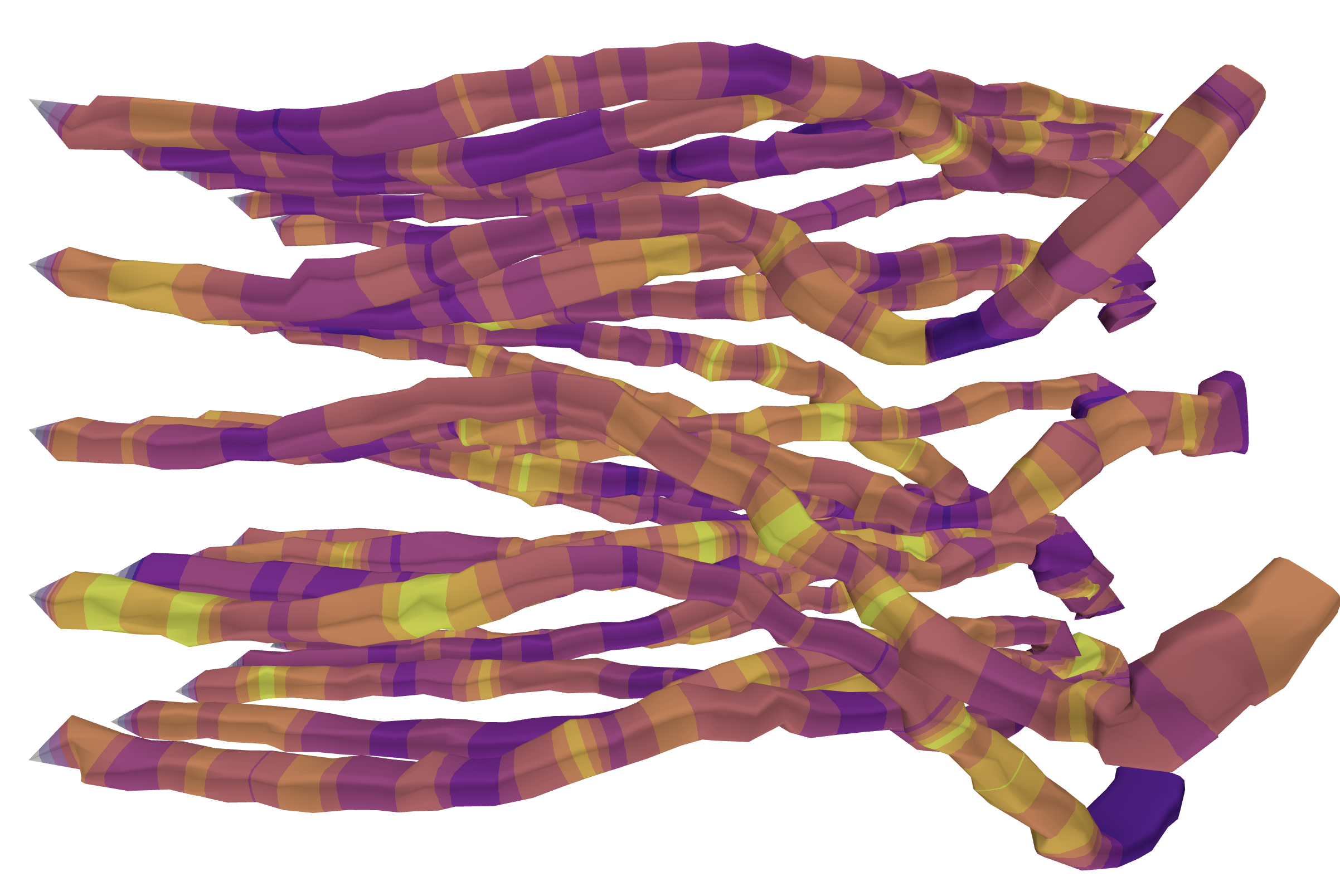}
        \caption{\swaglr=1e-2}
    \end{subfigure}
    \hfill
    \begin{subfigure}{0.3\textwidth}
        \centering
        \includegraphics[width=1\textwidth]{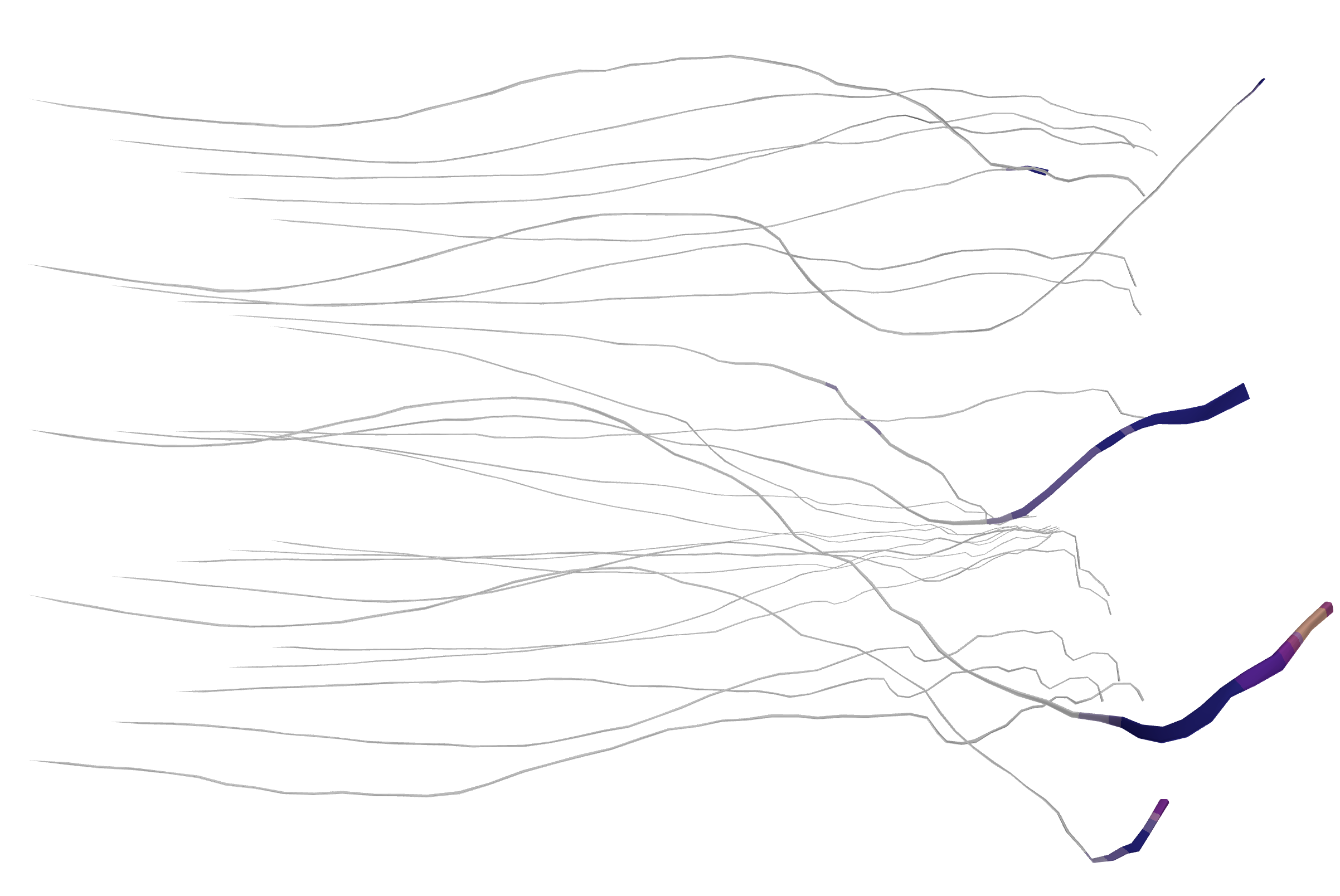}
        \caption{\swaglr=1e-4}
        \label{fig:synth_swag_ht_lr_1e-4}
    \end{subfigure}
    \hfill
    \begin{subfigure}{0.3\textwidth}
        \centering
        \includegraphics[width=1\textwidth]{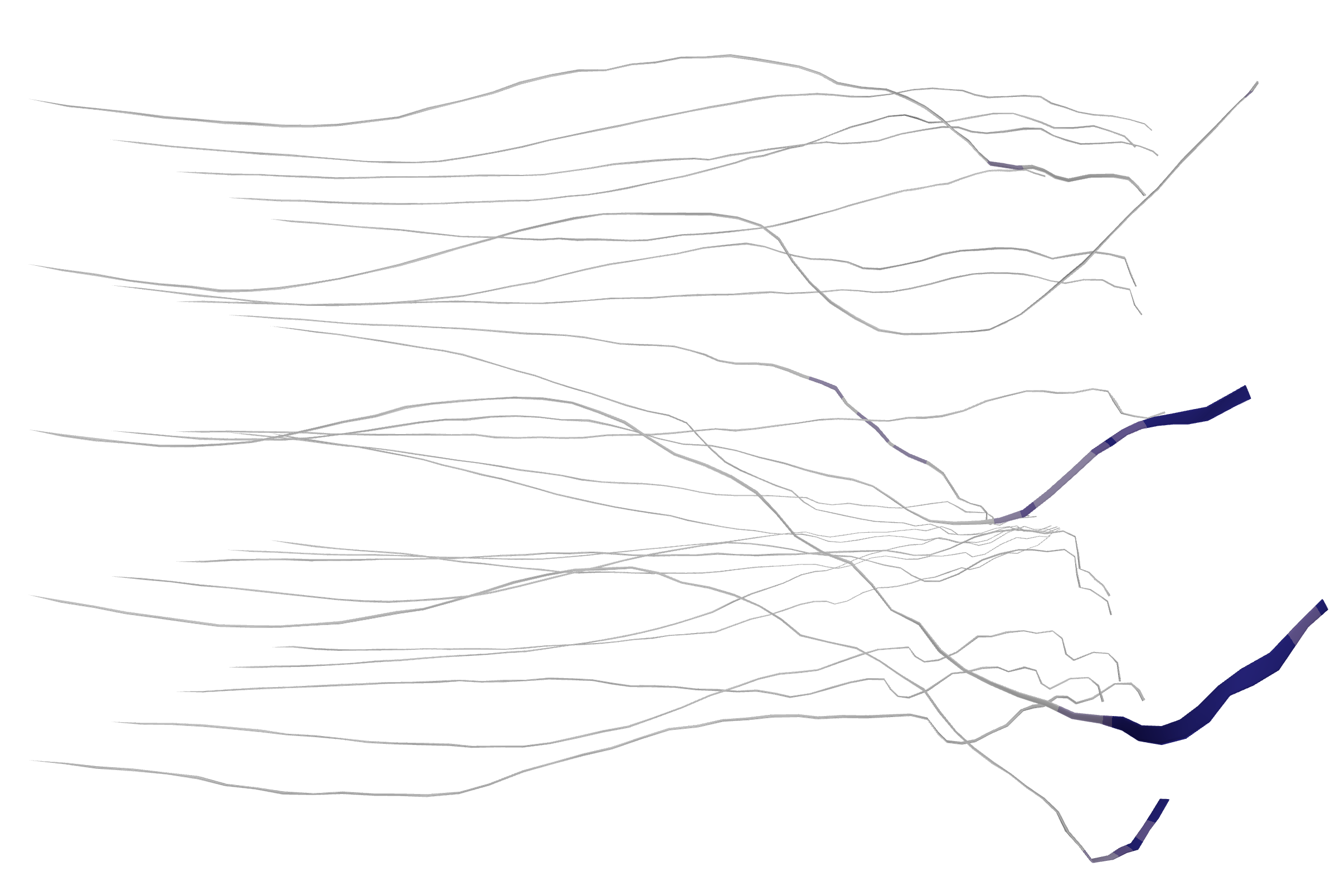}
        \caption{\swaglr=1e-8}
        \label{fig:synth_swag_ht_lr_1e-8}
    \end{subfigure}
    \\
    \begin{subfigure}{0.3\textwidth}
        \centering
        \includegraphics[width=1\textwidth]{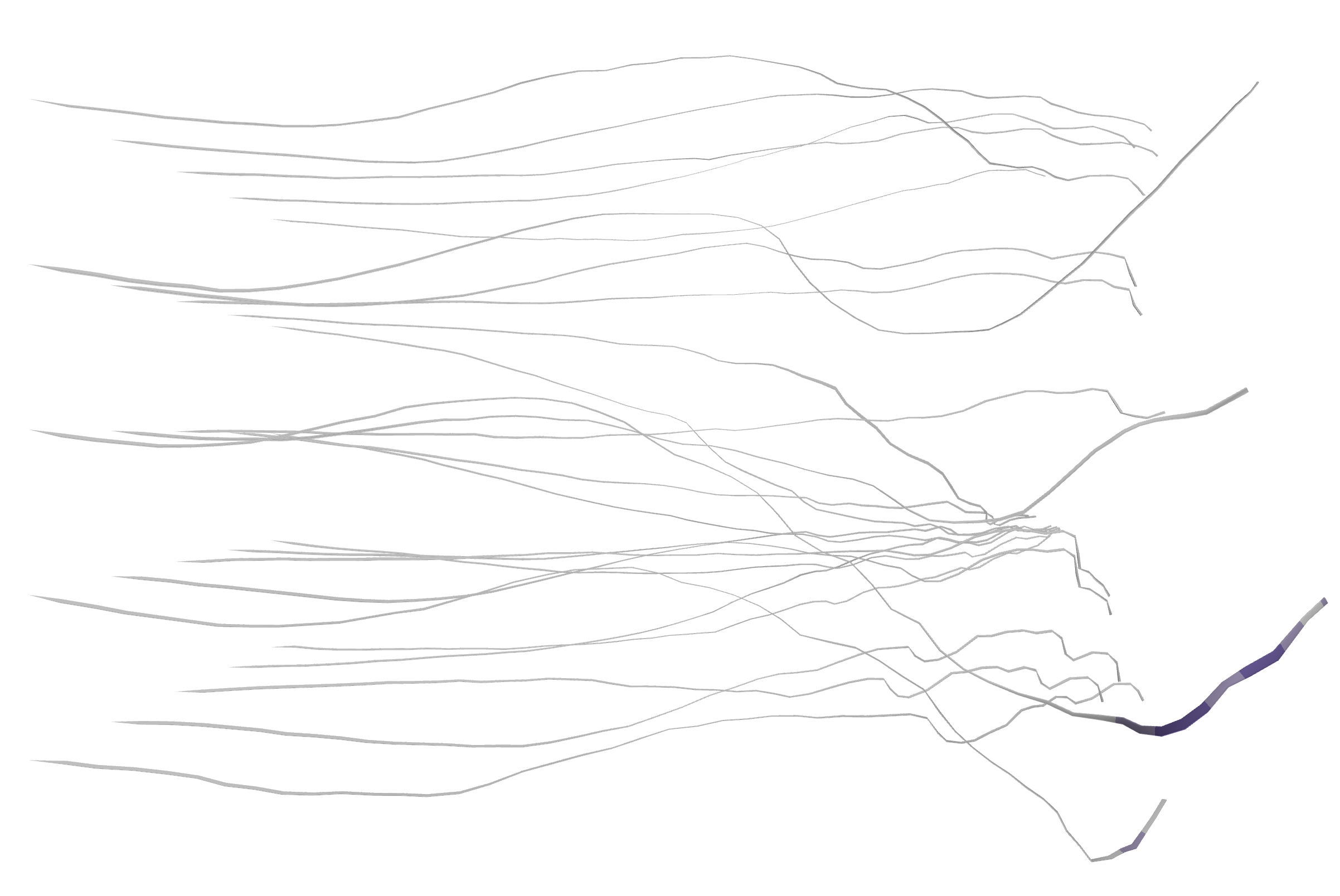}
        \caption{\nswagsamples=10, rank = 10}
        \label{fig:synth_swag_ht_samples_10}
    \end{subfigure}
    \hfill
    \begin{subfigure}{0.3\textwidth}
        \centering
        \includegraphics[width=1\textwidth]{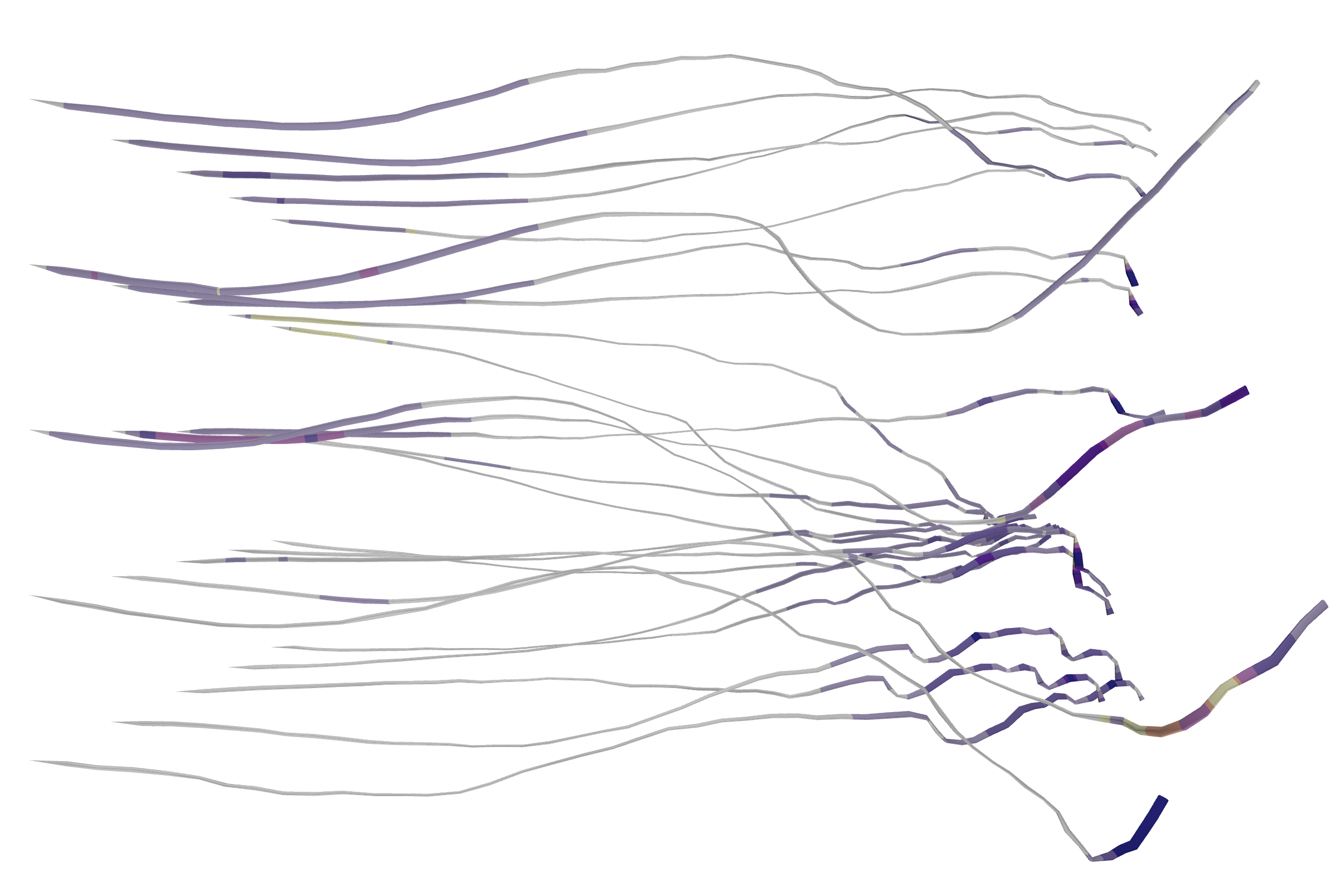}
        \caption{\nswagsamples=50, rank = 50}
    \end{subfigure}
    \hfill
    \begin{subfigure}{0.3\textwidth}
        \centering
        \includegraphics[width=1\textwidth]{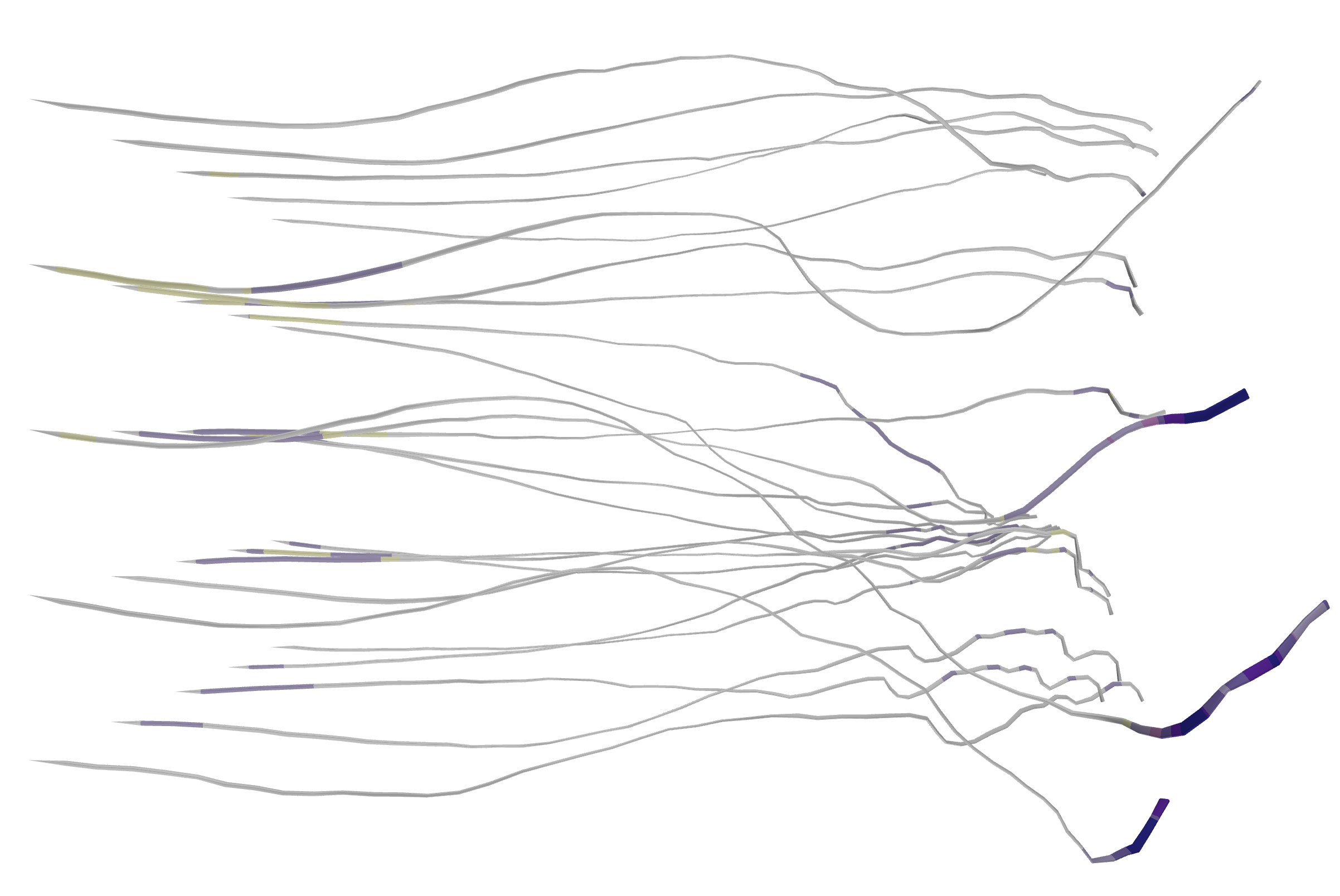}
        \caption{\nswagsamples=100, rank = 100}
    \end{subfigure}
    \\
    \begin{subfigure}{0.3\textwidth}
        \centering
        \includegraphics[width=1\textwidth]{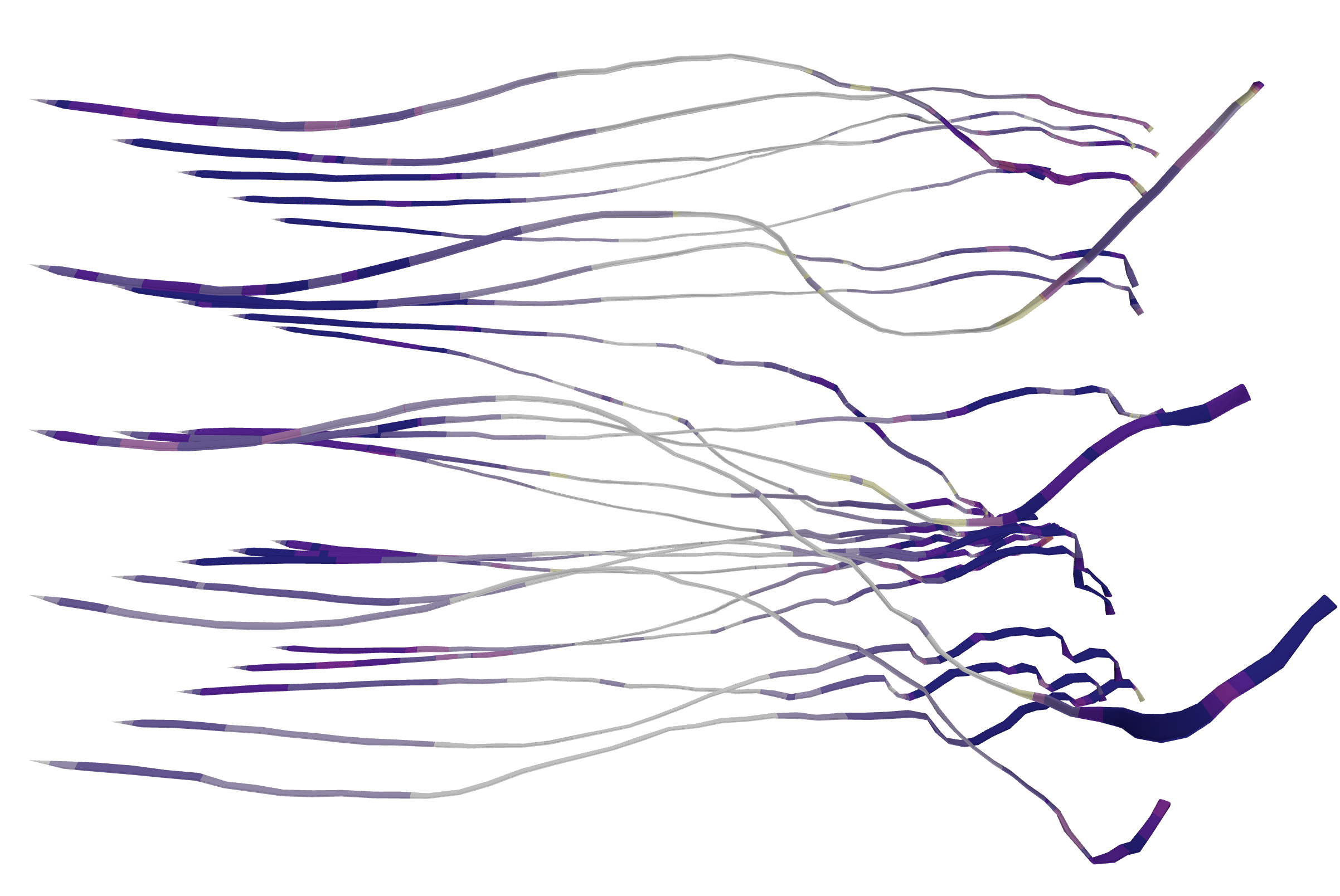}
        \caption{rank=10}
    \end{subfigure}
    \hfill
    \begin{subfigure}{0.3\textwidth}
        \centering
        \includegraphics[width=1\textwidth]{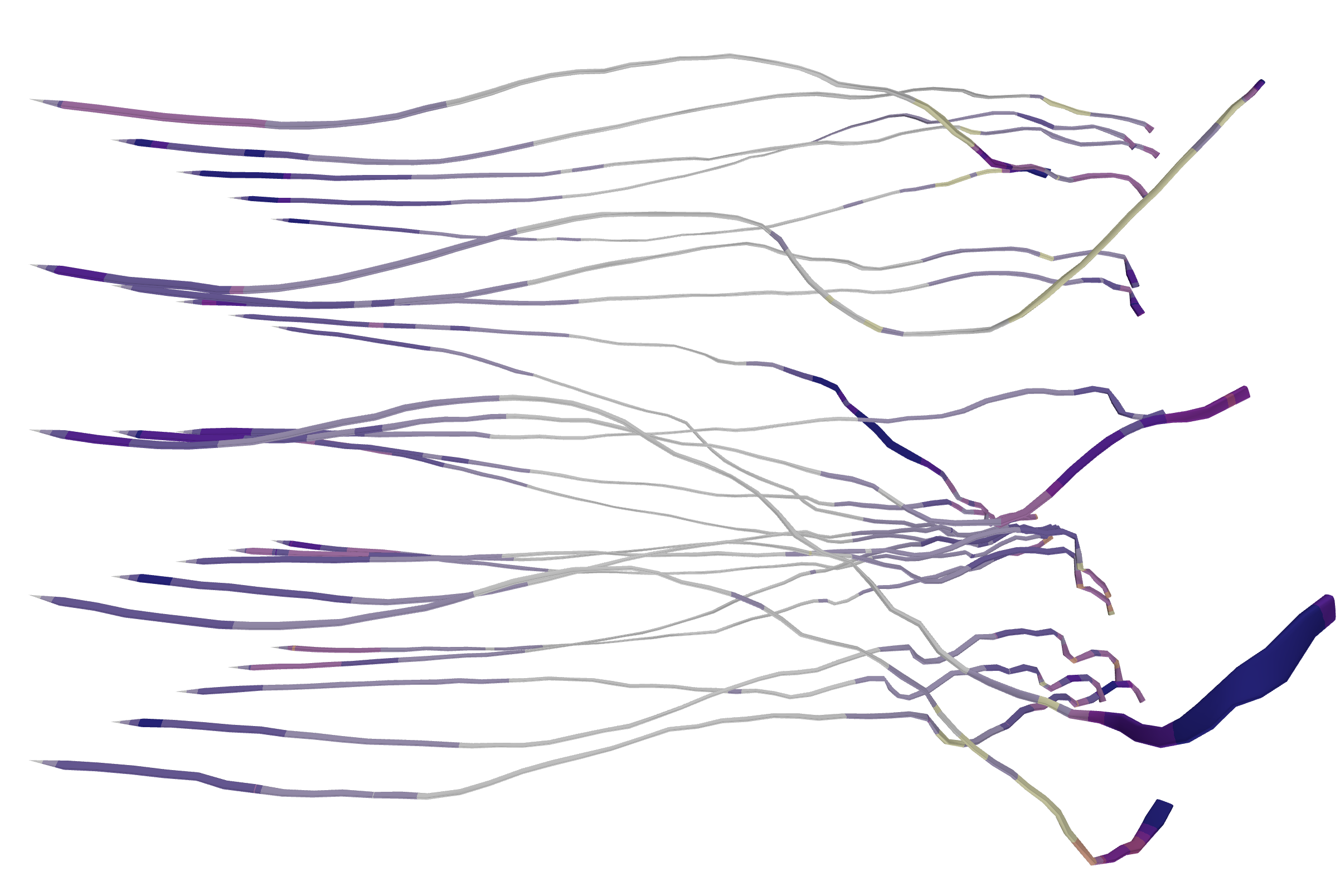}
        \caption{rank=50}
    \end{subfigure}
    \hfill
    \begin{subfigure}{0.3\textwidth}
        \centering
        \includegraphics[width=1\textwidth]{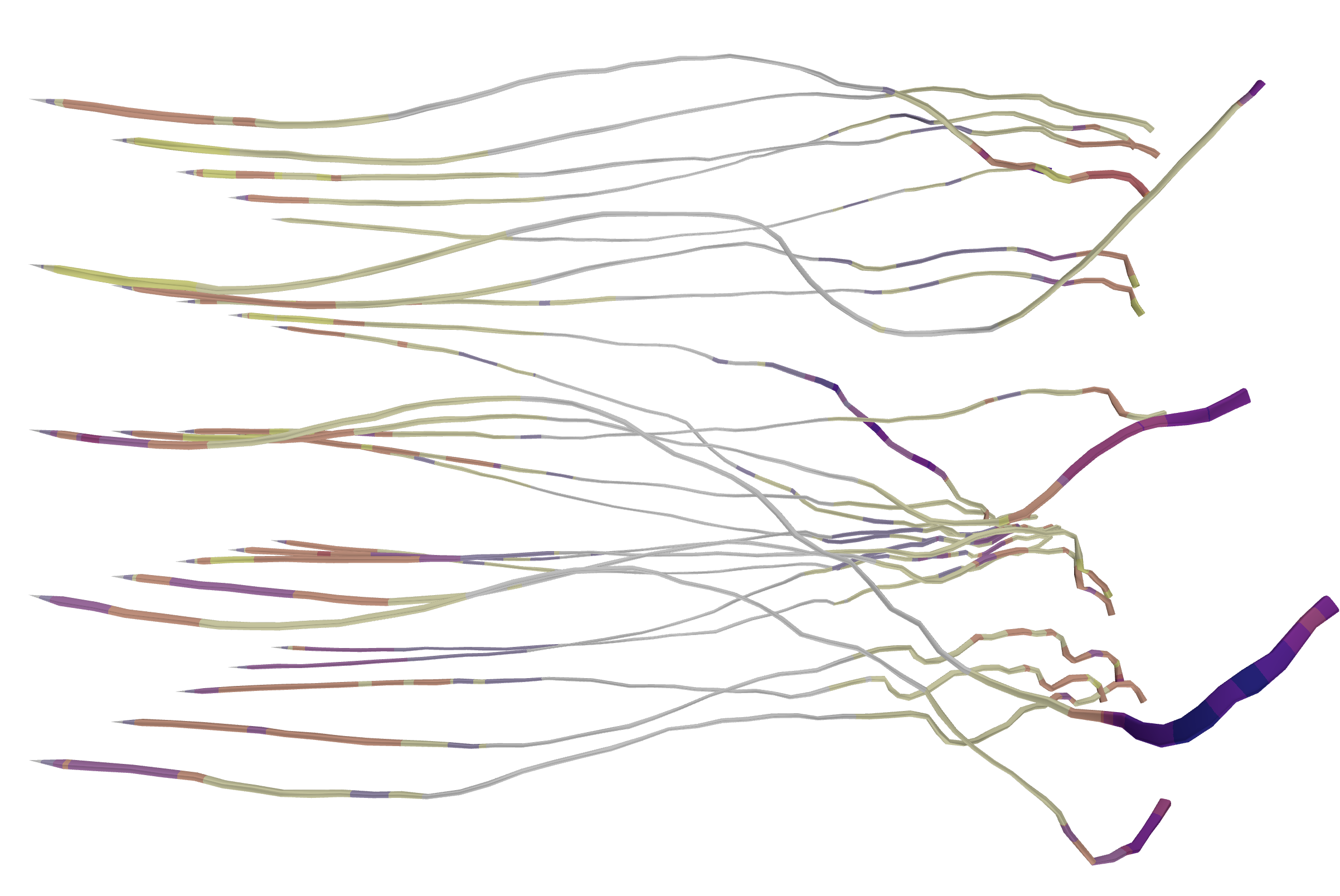}
        \caption{rank=1000}
    \end{subfigure}
    \caption{Hyperparameter test results of SWAG.}
    \label{fig:synth_swag_ht}
\end{figure*}

Here, we present a hyperparameter study for our \synth~model. In our implementation, we exposed five hyperparameters: \swaglr, \nswagsamples, rank, \sgdweightdecay, \sgdmomentum. 
The \swaglr~controls the step size of the SGD in SWAG training. Maddox et al.~\cite{maddox_2019_swag} recommend a high constant learning rate so that SGD explores the model's weight space instead of simply converging to a local minimum. 
The number of steps SWAG training takes is the \nswagsamples. It needs to be set to a sufficiently large number to explore the weight space. And it heavily impacts the SWAG training time.
Rank is the rank of the low-rank approximation of the covariance matrix of the multivariate Gaussian. The covariance matrix captures the correlation among the model parameters.
Maddox et al.~\cite{maddox_2019_swag} stated that weight decay and momentum need to be explicitly specified. Then, SWAG can be viewed as a Bayesian inference approximation because weight decay with momentum corresponds to the prior distribution of the model weights.

We start from the base hyperparameter used in \cref{fig:synth_swag}: \swaglr=$5e-4$, \nswagsamples=1000, rank=$100$, \sgdweightdecay=$1e-8$, \sgdmomentum=0.9. We tested the $\swaglr=[1e-2,1e-4,1e-8]$, $\nswagsamples=[10, 50, 100]$, $rank=[10, 50, 1000]$. For \nswagsamples, we also match the rank. The \sgdweightdecay~and \sgdmomentum~parameters are set to $1e-8$ and $0.9$, respectively. We omit exploring these two parameters and recommend setting them according to the optimal training parameters.

The hyperparameter exploration results are presented in \cref{fig:synth_swag_ht}. For the \swaglr~parameter, we recommend examining the global pattern of the network's predictions. That is, if uncertainty is globally high, we recommend reducing the \swaglr. In our test, we tested up to $1e-8$, and the SWAG samples still show reasonable uncertainty because we set \nswagsamples~sufficiently large. \nswagsamples~controls the number of steps SWAG takes in the model's weight space, and \swaglr~controls the step size. The combination of those should be large enough to explore the weight space. \cref{fig:synth_swag_ht_samples_10} shows the classical underexplored weight space, showing no uncertainty in the model's prediction. The \nswagsamples~parameter has a significant impact on the SWAG training time. In our experiments, the time it takes to train for $10$, $50$, $100$, and $1000$ steps is $1$ second, $3$ seconds, $6$ seconds, and $15$ seconds, respectively. We use a conservative setting of $1000$ in this study.

As shown in the third row of \cref{fig:synth_swag_ht}, the effect of the rank parameter is not straightforward. The rank parameter sets the rank of the low-rank approximation of the covariance matrix of the model parameters. The consequence of the correlation between model parameters on the final prediction is not immediately clear to us. If the model is sufficiently small, we recommend examining the singular values obtained from the singular value decomposition of the full-rank covariance matrix and selecting the smallest reasonable rank for the low-rank approximation. 
In our experiment across different datasets and models, we found that  $100$ is a sufficiently large number.
The rank also affects the SWAG training time and the time it takes to draw samples from the Gaussian. For ranks $10$, $50$, $500$, $1000$, the training takes $13$, $13$, $17$, $29$ seconds for $1000$ \nswagsamples. Drawing 50 samples from the fitted Gaussian takes $32$, $61$, $320$, and $700$ milliseconds, respectively.

Across all hyperparameter tests, we compute the \textit{uncertainty tube} using $25$ seeds and $50$ uncertainty samples, in addition to the original model's prediction. The model evaluation for the $25\times 51$ pathlines takes 90 milliseconds, and the computation of the $25$ uncertainty tubes takes approximately $350$ milliseconds.

\section{Results}
\label{sec:results}
\begin{figure*}[!htb]
    \begin{subfigure}[t]{1.0\columnwidth}
        \centering
        \includegraphics[width=1\textwidth]{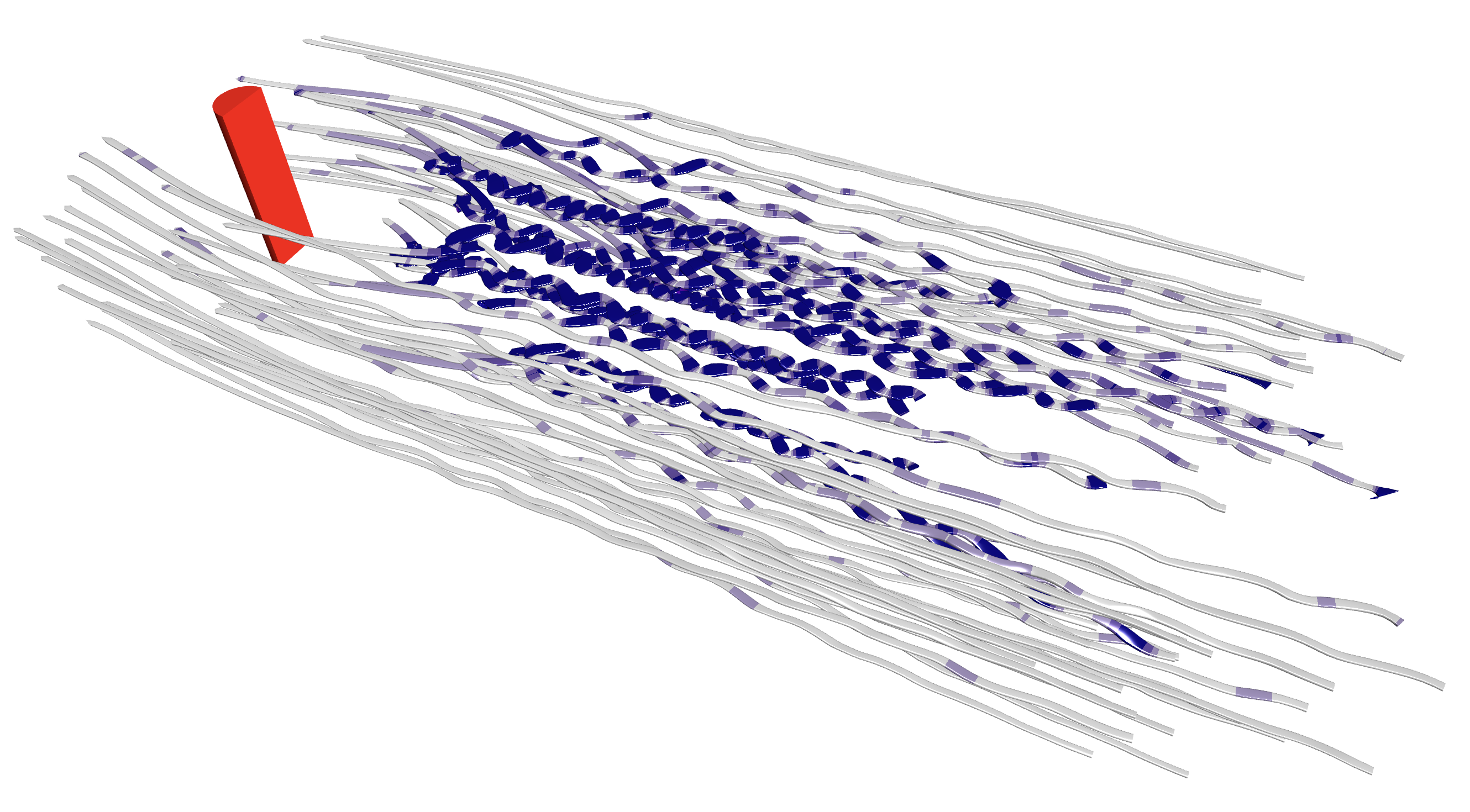}
        \caption{Uncertainty from a model trained according to Han et al.\cite{han2024interactive}.}
        \label{fig:halfcylinder}
    \end{subfigure}\hfill
    \begin{subfigure}[t]{1.0\columnwidth}
        \centering
        \includegraphics[width=1\textwidth]{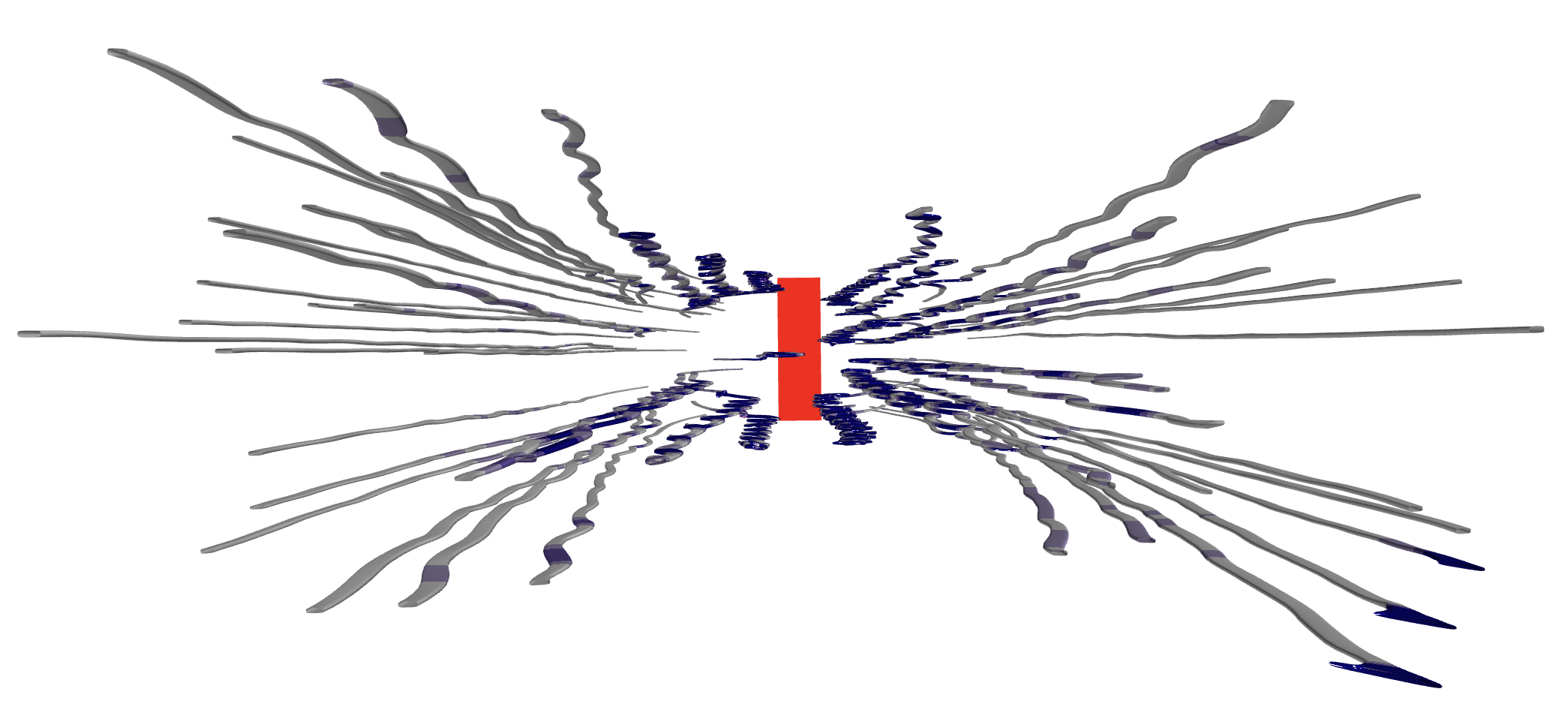}
        \caption{A view looking at yz-plane from +x.}
        \label{fig:halfcylinder_xview}
    \end{subfigure}
    \\
    \begin{subfigure}[t]{1.0\columnwidth}
        \centering
        \includegraphics[width=1\textwidth]{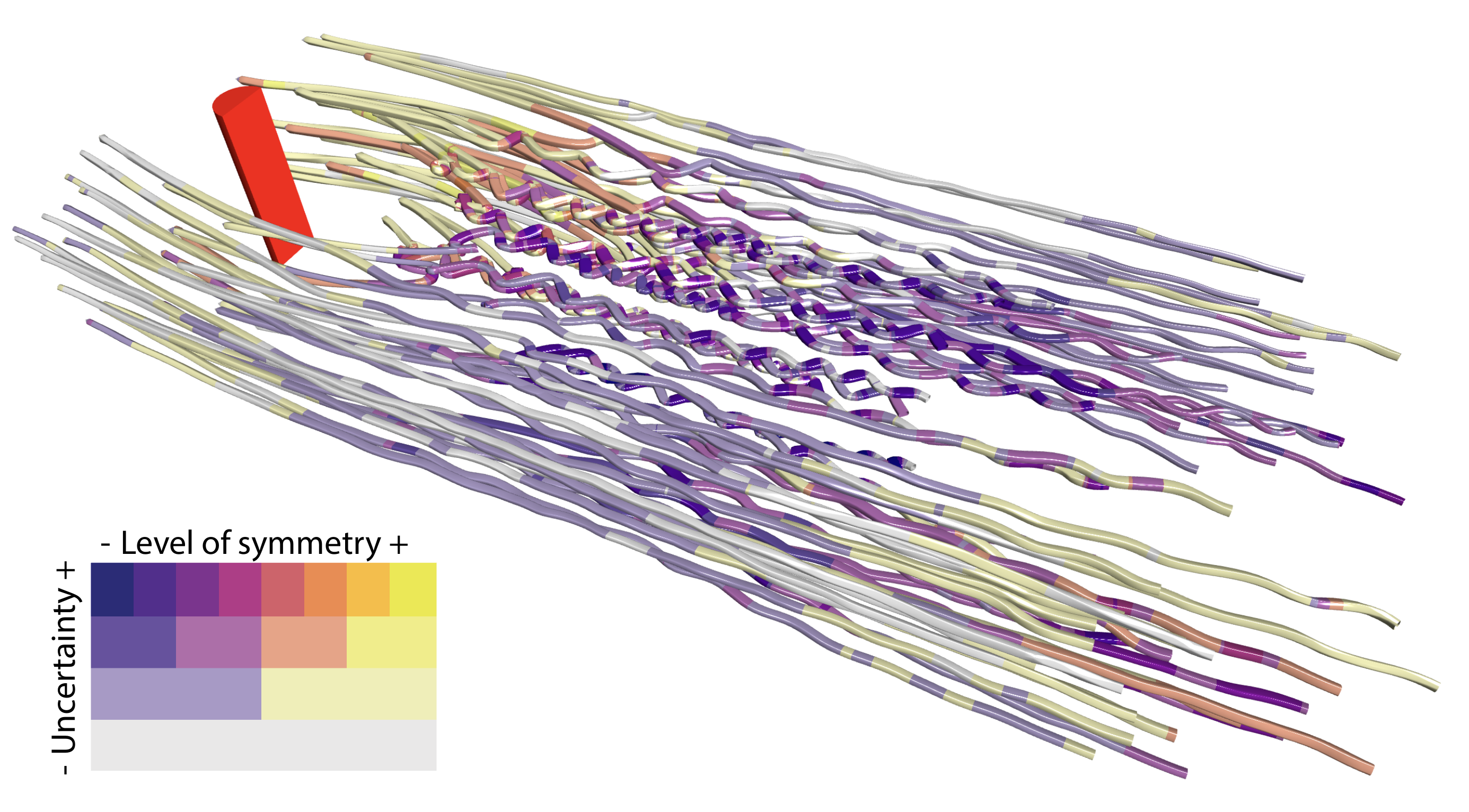}
        \caption{Uncertainty from a model trained with spatially uniform scaling.}
        \label{fig:halfcylinder_rescaled}
    \end{subfigure}\hfill
    \begin{subfigure}[t]{1.0\columnwidth}
        \centering
        \includegraphics[width=1\textwidth]{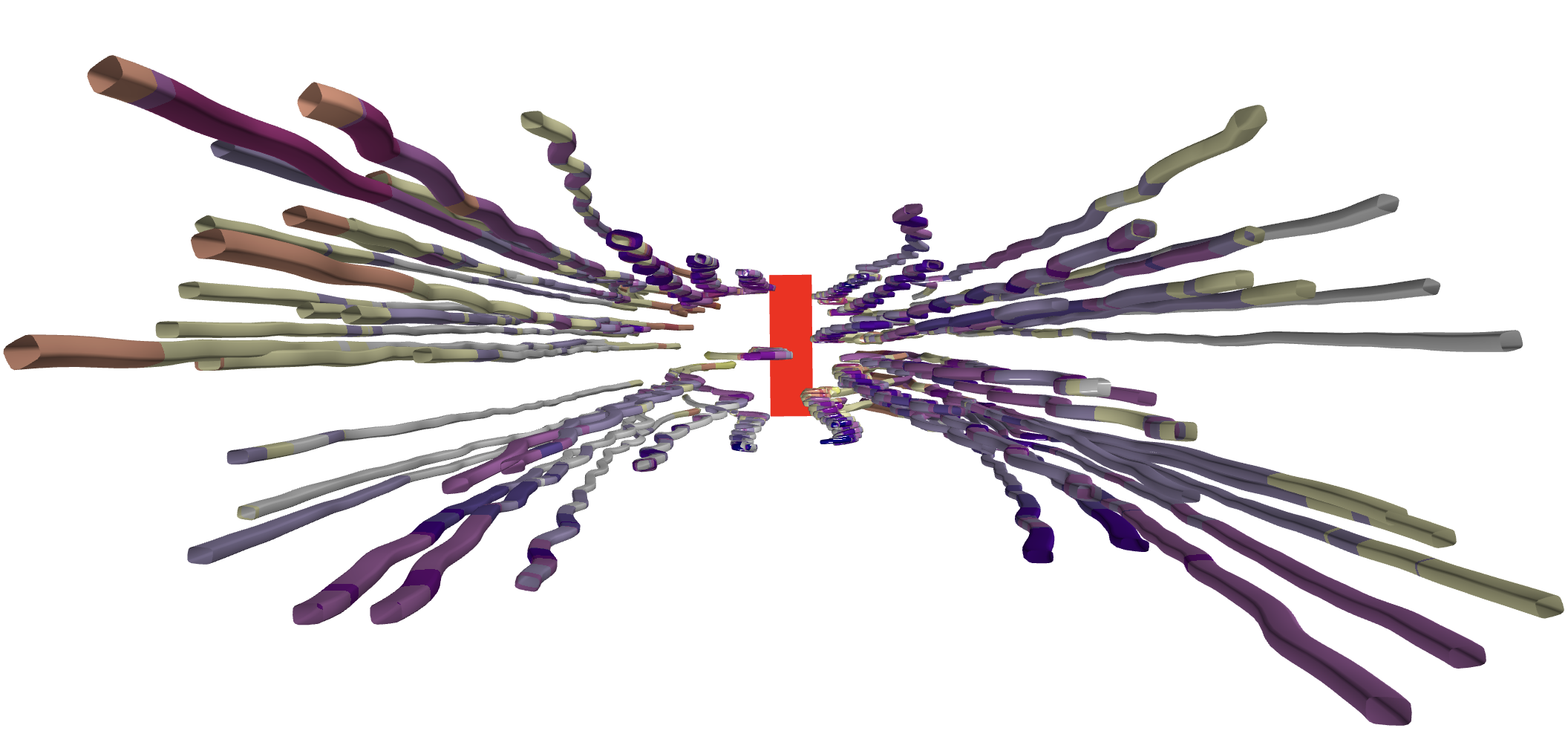}
        \caption{A view looking at yz-plane from +x.}
        \label{fig:halfcylinder_rescaled_xview}
    \end{subfigure}
    \caption{Uncertainty Tube visualization of \halfcylinder~dataset.}
    \label{fig:halfcylinders}
\end{figure*}
\subsection{Tornado}
\label{subsec:tornado}
This example utilizes flow maps generated from a synthetic \textbf{tornado} vector field dataset in Güther et al.~\cite{Gunther2013}. We employ a flow map NN with four encoder layers, a $512$ latent vector, and six decoder layers. The NN is trained on $131072=2^{17}$ trajectories integrated over $100$ time steps with $\delta =0.1$ and using Sobol seeds within subdomain $[-5,5]\times[-5,5]\times[-10,10]$.  Our goal is to compare different visualization techniques for visualizing the model uncertainty calculated using the SWAG method. 

\autoref{fig:teaser} compares the spaghetti plot of ensemble members, the circular tube visualization, and the \textit{uncertainty tube} for representing the model uncertainty. The ensemble visualization in \cref{fig:teaser}.a leads to visual clutter that hinders the interpretation of the flow patterns. Additionally, the ensemble trajectories do not intuitively convey the model's uncertainty. The circular tube visualization in \cref{fig:teaser}.b reduces the clutter and provides a better visual summary of the uncertainty compared to the spaghetti plot. However, it incorrectly assumes a symmetric distribution of trajectories around the mean. The \textit{uncertainty tube} described in \cref{subsec:uncertainty-tube} encodes the variation direction using the major and minor axes of superellipses, as shown in \cref{fig:teaser}.c. Our \textit{uncertainty tube} enhances the distinction between the major and minor directions of variation and the visualization of how the direction changes along the trajectory integration, as depicted in boxes shown in \cref{fig:teaser}. \change{The \textit{uncertainty tube} in the right box provides a more detailed representation, where changes in orientation and twisting along the trajectory are distinctly visible. In contrast, the circular tube on the left fails to capture or convey these variations. In cases of subtle changes along the trajectory, the color palette further enhances the distinction between low and high uncertainty, and between symmetric and nonsymmetric uncertainty.} Overall, the \textit{uncertainty tube} yields superior results in estimating and visualizing model uncertainty compared to the other approaches.

\subsection{Half cylinder}

The \halfcylinder~dataset is a time-varying flow field simulating flow over a half cylinder towards the positive x-direction. Visualizing asymmetric uncertainty enabled us to train a more accurate model.

In Han et al. \cite{han2024interactive}, the data is mapped into a $[-1, 1]^3$ space by rescaling the bounding box of the data. In \cref{fig:halfcylinder}, we notice that all the \textit{uncertainty tubes} appear flat in the $y$-direction, as shown in \cref{fig:halfcylinder_xview}.
We suspect the flatness is caused by nonspatially uniform scaling of the data.
We rescaled the data according to the actual ratio in the domain, so $[-1, 1]^3$ represents a cube of size  $8^3$ in the domain. Although the training box $[-1, 1]^3$ now contains empty spaces, spatially uniform scaling has reduced the evaluation error from 0.0058 to 0.0047. The evaluation error represents the absolute difference between the predicted location and the true location of the validation dataset in the original domain. We ran multiple random seeds to compare different training processes and consistently observed a reduction in the evaluation error. 
We also noticed a similar improvement by applying spatially uniform scaling of the \textbf{Hurricane} dataset used in Han et al.~\cite{han2024interactive}. However, the generalizability of this finding to other MLP-based networks with sine activation is beyond the scope of this paper.

% We suspect that a multilayer perceptron network with sine activation may work better when the data is mapped to $[-1, 1]^d$ (centered at [0, 0, 0]) in a spatially uniform manner. But we did not verify this hypothesis in this paper.

Another interesting effect is that, despite improved model quality, the amount of uncertainty (measured by the maximum eigenvalue) increased. This is illustrated by \cref{fig:halfcylinder} showing more gray than \cref{fig:halfcylinder_rescaled}. Throughout our experiment, we consistently observed a mismatch between the amount of error and the model's uncertainty estimation. This mismatch means that extra caution is required when interpreting the results of uncertainty estimation, as they do not always reflect the actual model error but the model's confidence in its prediction. 

\section{Discussion}
This paper introduced the \textit{uncertainty tube}, a novel and computationally efficient visualization method for representing prediction uncertainty in neural network-derived particle trajectories. We design and implement a superelliptical tube that uniquely captures and intuitively conveys asymmetric uncertainty, thereby overcoming the limitations of conventional methods that typically assume symmetric uncertainty bounds. 
By integrating well-established uncertainty quantification techniques, including Deep Ensembles, MC Dropout, and SWAG, we demonstrated that the \textit{uncertainty tube} significantly improves the representation of asymmetric uncertainty compared to the circular tube. Moreover, its rectangular design distinctly shows the uncertainty orientation and its evolution along the pathline more effectively than the circular and elliptical tubes. 
Our VSUP-inspired color map further helps distinguish different types of uncertainty when visualizing 3D geometries. In addition, we demonstrated one use case where we utilize asymmetric uncertainty to enhance training. We hope to explore additional ways to utilize uncertainty information to better understand the data, training, and models in the future. 

The \textit{uncertainty tube} visualization of the trajectory has some limitations. For example, when constructing the \textit{uncertainty tube}, we projected the point at each step onto the plane orthogonal to the mean trajectory. This process eliminates the \change{uncertainty along the mean trajectory}. We could use color or texture to represent that type of uncertainty, or use local superquadric glyphs~\cite{Thomas2010} to characterize the variation of all directions. Here, we focus on developing a visual representation that highlights the asymmetric nature of uncertainty in NN-based trajectories. A user-based study is necessary to evaluate the effectiveness and expressiveness of visual encoding for future research. 
%Extending this approach to handle bifurcations will further generalize its applicability to flow uncertainty that exhibits such behavior. 

An important next step involves investigating more comprehensive uncertainty quantification methods, such as fully modeled Bayesian networks, especially to investigate how uncertainty is propagated through the data analysis pipeline. The way we utilized the three UQ methods in this paper assumes no uncertainty in the training data, which is rarely true in real-world applications.

\acknowledgments{
This work was partially supported by the Intel OneAPI CoE, the Intel Graphics and Visualization Institutes of XeLLENCE, and the DOE Ab-initio Visualization for Innovative Science (AIVIS) grant 2428225.}
%-------------------------------------------------------------------------
% bibtex
\bibliographystyle{abbrv-doi}
\bibliography{main}       

% biblatex with biber
% \printbibliography                

\end{document}